\documentclass[10pt,twocolumn,letterpaper]{article}

 \usepackage{wacv}              

\usepackage[accsupp]{axessibility}  

\usepackage{algorithm}
\usepackage{algorithmic}

\definecolor{wacvblue}{rgb}{0.21,0.49,0.74}
\usepackage[pagebackref,breaklinks,colorlinks,allcolors=wacvblue]{hyperref}

\def\wacvPaperID{2035} 
\def\confName{WACV}
\def\confYear{2027}

\title{FLASH: A “Generate Once, Synthesize Many” Framework for Synthetic Anomaly Generation in Industrial Anomaly Detection }

\author{
    Abhay Kumar Das$^1$
    \and
    Rajesh Gangireddy$^2$
    \and
    Ashwin Vaidya$^2$
    \and
    Samet Akcay$^2$
    \and
    $^1$Silicon University, Bhubaneswar, India
    \and
    $ ^{2}$Intel \\
    {\tt\small dasabhay.jsr@gmail.com}, 
    {\tt\small \{rajesh.gangireddy, ashwin.vaidya, samet.akcay\}@intel.com}
}

\begin{document}
\raggedbottom
\maketitle
\begin{abstract}
Synthetic anomaly generation helps expand industrial anomaly datasets when real defects are scarce or unavailable. Existing approaches lie at two extremes: procedural approaches are fast but struggle to represent complex anomalies, while generative approaches produce diverse defects but require costly per-sample generation.
We present FLASH, a framework that decouples defect generation from anomaly synthesis under a ``generate once, synthesize many'' paradigm. Given only normal images, FLASH uses vision-language model (VLM) guidance and an image-generation model to produce a small set of defect images, from which it extracts, validates, and banks reusable defect patches. For synthesis of anomalous images, Object Boundary Suppression (OBS) first identifies the probable foreground object-aware region of the host image, while Multi-Resolution Spectral Pyramid (MRSP) noise generates diverse, size-controllable masks that determine the defect location and spatial extent. It then composes a large and diverse synthetic anomalous image set by localizing the defect region, sampling size-controllable placement masks and seamlessly blending retrieved defects onto new defect-free images without further need for image generation.
Experiments on the MVTec AD 2 dataset show that FLASH-generated anomalies nearly close the
calibration gap on real defects, reaching 78.1\% image-level F1 against an
83.6\% real-anomaly reference and providing the most consistent calibration transfer across detectors among procedural and generative alternatives. Moreover, FLASH provides an estimated 13.8$\times$ image-generation speedup over AnoStyler under the stated timing assumptions. Code and configuration files are available at {\url{https://github.com/AbhayKumarDas/flash}}
\end{abstract}

\vspace{-12pt}
\section{Introduction}
\label{sec:intro}

In manufacturing, production lines predominantly yield defect-free products, while anomalous samples are rare, diverse, and expensive to collect and annotate \cite{ref11}. This makes synthetic anomaly generation an attractive alternative for constructing anomalous data from normal images \cite{ref170, ref165}. Existing approaches occupy two extremes: computationally faster procedural and feature-based methods, and slower but contextually coherent generative methods. The first set of methods is limited in its ability to represent complex and semantically meaningful defects \cite{ref89,ref97}, whereas the second, though closer to the target domain, requires substantially more time and computation per generated sample \cite{ref67, ref78}. We therefore aim to develop a framework that lies between these two extremes: retaining the fast generation and computational efficiency of procedural
synthesis while obtaining diverse synthetic anomalies through limited generative inference.

We introduce FLASH, a framework designed around this trade-off. It generates a small set of defect instances once and reuses them to synthesize a substantially larger anomaly dataset, thereby reducing generation time while operating without any real defect references. The resulting synthetic anomalies are context-specific, thus serving as a viable surrogate in the absence of real defects for downstream anomaly detection (AD), including decision-threshold calibration \cite{ref174}. FLASH achieves this through a five-stage approach. Its factorized design
allows a limited set of generated defects to be reused across normal images,
balancing generation speed, computational efficiency, and synthetic anomaly
utility. We evaluate FLASH on MVTec AD 2~\cite{ref60}, a challenging benchmark
with diverse conditions and anomaly types beyond conventional structural
defects, including subtle foreign contamination. Such anomalies challenge
procedural synthesis, where predefined noise or feature perturbations have
limited capacity to model the appearance, context, and spatial characteristics
of externally introduced materials, making MVTec AD 2 a stringent test of
whether reference-free synthesis can move beyond simple procedural defects.

\section{Related Work}
\label{sec:formatting}

Synthetic anomaly generation encompasses procedural methods that prioritize
efficient synthesis and generative approaches that emphasize contextual
fidelity, forming a persistent procedural--generative trade-off.

\subsection{Procedural Anomaly Generation }

Procedural methods establish a simple route to reference-free anomaly synthesis by transforming normal images through patch relocation \cite{ref89}, blending \cite{ref152}, and Poisson cloning \cite{ref138}. Noise-based formulations extended this principle to scalable defect formation using Perlin masks and sampled textures \cite{ref170, ref114}, with subsequent work improving object constraints \cite{ref166} and perturbation subtlety \cite{ref26}. Their low computational cost makes them attractive for large-scale synthesis; however, the generated defect space remains tied to predefined perturbations or borrowed textures, limiting their ability to represent complex context-dependent anomalies such as those arising from real manufacturing process errors.

\subsection{Generative Anomaly Generation }

  Generative models address the above limitations by broadening the diversity of defects that can be represented. Generative adversarial network (GAN) approaches introduced defective counterparts
  \cite{ref106}, controllable synthesis \cite{ref171}, and defect transfer from
  limited anomalous samples \cite{ref40}. Diffusion-based approaches \cite{ref129} further enabled anomaly embeddings \cite{ref67}, joint image-mask generation
  \cite{ref78}, controllable anomaly strength \cite{ref177}, inpainting
  \cite{ref146}, cross-category transfer \cite{ref140}, and spatial conditioning
  \cite{ref54}. Vision-language models (VLMs) subsequently added contextual reasoning to
  anomaly analysis \cite{ref72, ref182, ref53} and synthetic defect generation
  \cite{ref86, ref77, ref66}. These advances broaden the range of defects that
  can be generated, but generative inference typically must be repeated for each
  synthesized sample, making the process slow and computationally expensive.
  Several approaches also rely on real anomalous references \cite{ref36}.

\subsection{Industrial AD Benchmarks and Detectors}
The practical value of synthetic anomalies depends on whether they provide useful anomalous evidence for downstream industrial anomaly detection. Most synthetic anomaly generation methods are benchmarked on MVTec AD \cite{ref11} and VisA \cite{ref184}. Performance on these datasets does not necessarily reveal how synthetic anomalies transfer to more challenging inspection conditions. MVTec AD 2 \cite{ref60} includes transparent objects, challenging illumination, and foreign contamination. We use it to evaluate synthetic anomalies beyond basic procedural perturbations.

Downstream anomaly detection methods span several architectural paradigms, including embedding-based (PaDiM \cite{ref37}, PatchCore \cite{ref131}), feature-mapping (SimpleNet \cite{ref97}), student-teacher networks (EfficientAD \cite{ref10}), reconstruction-based frameworks (Reverse Distillation \cite{ref38}), and modern foundation-model approaches (Dinomaly \cite{dinomaly_cvpr}, AnomalyDINO \cite{ref193}, SuperADD~\cite{ref190}). However, existing benchmarks predominantly evaluate models using decision thresholds calibrated on real anomalies, an assumption that breaks down in practical zero-shot or unsupervised industrial settings where real anomalies are unavailable prior to deployment.

\section{Methodology}
\label{sec:methodology}

\begin{figure*}[t]
  \centering
  \includegraphics[width=\textwidth]{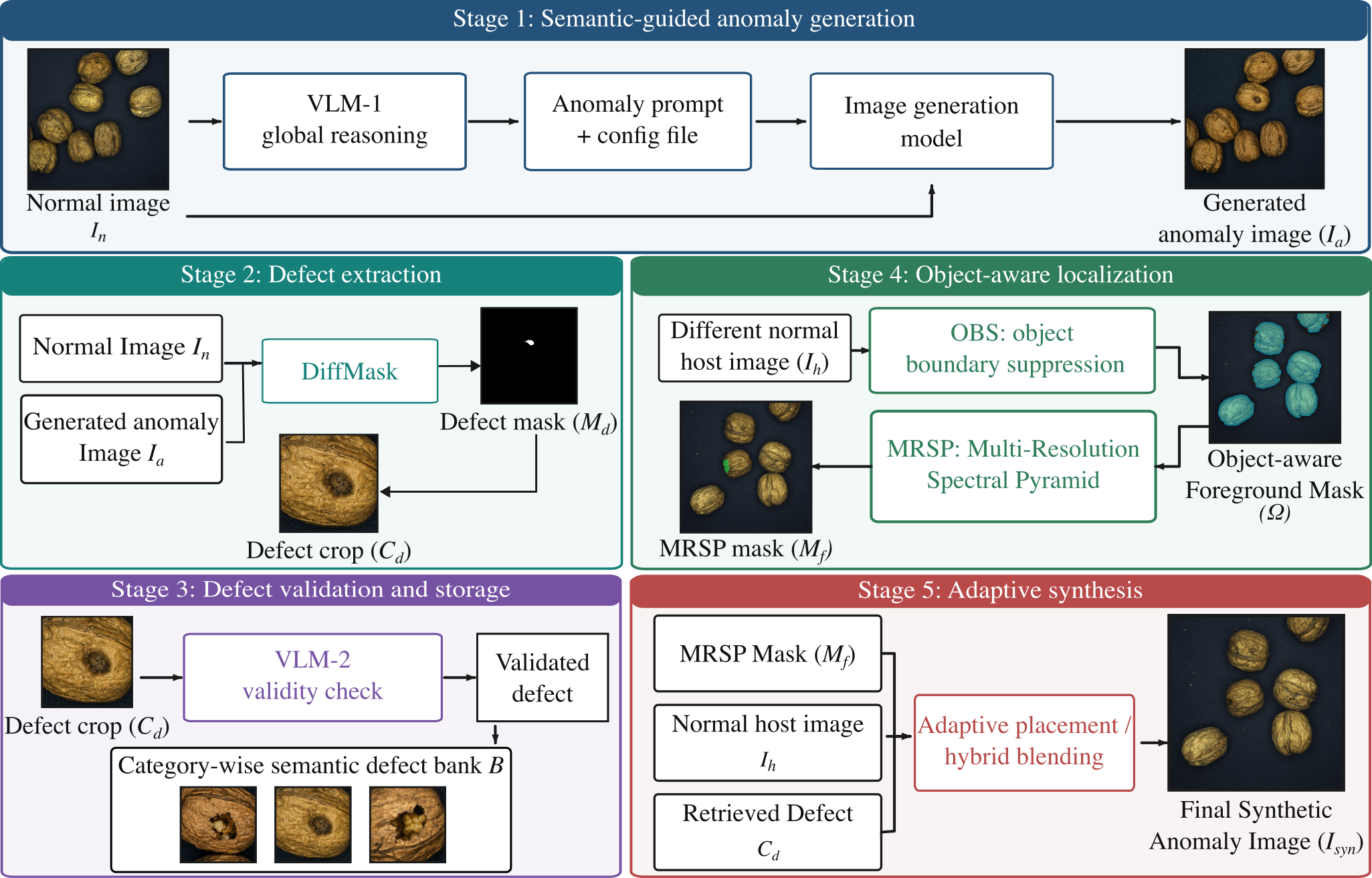}
  \caption{Five-stage FLASH architecture. Stages~1--3 generate, extract, validate, and store reusable defects, while Stages~4--5 localize and synthesize retrieved defects onto separate normal host images.}
  \label{fig:flash_architecture}
\end{figure*}

Industrial anomaly detection focuses on normal data as defective samples are rare, unevenly distributed across defect types, and expensive to collect and annotate at the pixel level \cite{ref11}. FLASH addresses this in its strictest form: given only normal images from a category, it generates anomalous images with pixel-accurate segmentation labels without using any real defect image, defect mask, or human annotation. Throughout the pipeline, $I_n$ denotes the Normal Image, $I_a$ the Generated Anomaly Image, $M_d$ the Defect Mask, $C_d$ the Defect Crop, $B$ the Category-wise Semantic Defect Bank, $I_h$ the Different Normal Host Image, $\Omega$ the Object-Aware region, $M_f$ the MRSP placement mask, $I_{syn}$ the Final Synthetic Anomaly Image, and $M_{gt}$ its corresponding ground-truth mask.

Existing pipelines are predominantly mask-first: a defect mask is defined before the defect is generated, after which the model is required to conform its output to the prescribed region \cite{ref166}. Diffusion-based approaches follow this formulation by conditioning image generation on predefined defect regions \cite{ref146}. This creates a mask-drift problem: the generated defect may not fully occupy the prescribed mask or may extend beyond it, causing disagreement between the anomaly and its label.
Unlike other approaches \cite{ref54}, FLASH generates the anomalous frame first and then recovers the actual Defect Mask $M_d$ from the generated image $I_a$. This mask can then be transferred onto a different Normal Host Image $I_h$ and placed within the MRSP mask $M_f$, where its location and spatial extent can be varied without distorting the recovered defect.

\Cref{fig:flash_architecture} presents the five-stage FLASH architecture. Stage~1 uses VLM reasoning about the entire image and a Config File to guide the image-generation model in producing $I_a$ from $I_n$. Stage~2 recovers $M_d$ using DiffMask and extracts the Defect Crop $C_d$, while Stage~3 validates the crop and stores it in the defect bank $B$. Stage~4 processes the Different Normal Host Image $I_h$ using OBS and MRSP to obtain the Object-Aware region $\Omega$ and placement mask $M_f$. Stage~5 retrieves a defect from $B$ and adaptively blends it within $M_f$ to produce $I_{syn}$ and $M_{gt}$ from $I_h$. Thus, $I_n$ serves as the donor for defect creation, while $I_h$ serves as the host for subsequent synthesis; the detailed operations of each stage are described in the following sections.

\begin{figure}[ht]
  \raggedright
  \includegraphics[width=0.99\columnwidth]{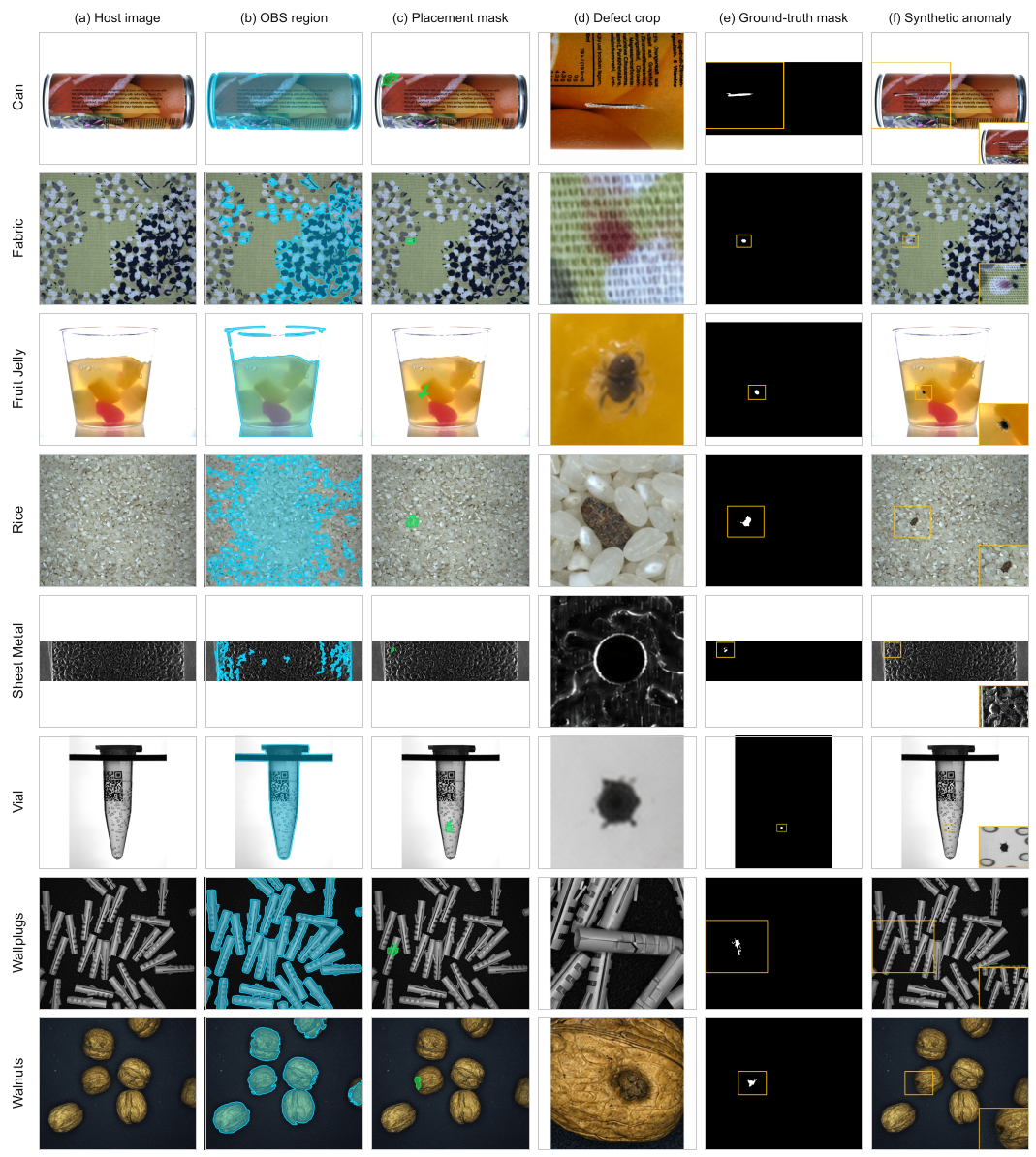}
    \caption{Qualitative results across all eight MVTec AD 2 categories, illustrating the end-to-end FLASH pipeline.}
  \label{fig:gallery_methology}
\end{figure}

Stages~1--3 are executed once per category to construct the defect bank, while Stages~4--5 execute for each output image. Thus, the image-generation model is not invoked for every synthetic sample, unlike fully generative pipelines whose generation cost scales with corpus size \cite{ref146}. The per-image generation instead relies on lightweight placement, harmonization and blending, allowing FLASH to retain the efficiency of procedural synthesis while using limited generative inference to obtain diverse contextual defects. The resulting defect bank can then be reused across normal host images, with the MRSP mask providing intra-image variation for additional diversity. An overview of the complete FLASH pipeline across all eight MVTec AD 2 categories, showing host images, OBS regions, placement masks, extracted defects, ground-truth masks, and final synthetic anomalies, is shown in~\Cref{fig:gallery_methology}.

\subsection{Semantic-Guided Anomaly Generation}
\label{sec:semantic_guided_anomaly_generation}

Starting from a Normal Image $I_n$, the first challenge is to specify what anomaly should be introduced before passing the image to the image-generation\footnote{OpenAI ChatGPT Images 2.5 \cite{ref185}} model. Rather than manually defining this prompt, we use VLM-1\footnote{Qwen2.5-VL-7B-Instruct \cite{ref9}} to reason globally over $I_n$, identify a plausible defect type and its material context, and generate an Anomaly Prompt. This prompt is augmented by a Config File that provides dataset context, category-wise defect types, and hard generation constraints. It requests photometric preservation with respect to $I_n$, including lighting, shadows, reflections, and color distribution, and prohibits changes to exposure, contrast, or white balance. As shown in~\Cref{fig:flash_architecture}, the Anomaly Prompt, Config File, and $I_n$ are provided to the image-generation model to produce $I_a$. Thus, VLM-1 provides the semantic specification of the anomaly, while the Config File provides dataset-specific context and generation constraints. Since the image-generation model regenerates the entire image with an anomaly, it may exhibit scale, spatial-offset, and pixel-level correspondence differences from $I_n$, motivating the Defect Extraction stage.

\subsection{Defect Extraction}
\label{sec:defect_extraction}

DiffMask frames the difference between the Normal Image $I_n$ and the
Generated Anomaly Image $I_a$ as a signal-recovery problem in which alignment, illumination, and reconstruction noise act as nuisance variation, while the introduced defect constitutes the signal. The two images are first registered using their edge maps to establish structural correspondence, followed by photometric adjustment.

\begin{figure}[ht]
  \raggedright
  \includegraphics[width=0.99\columnwidth]{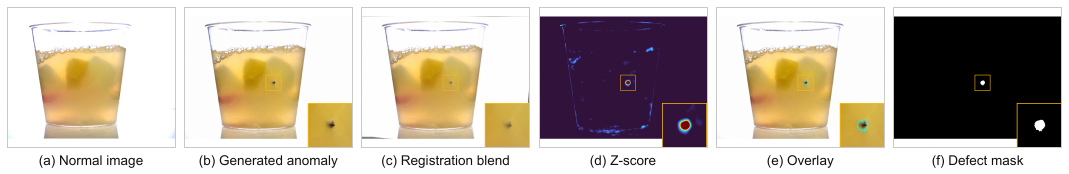}
  \caption{DiffMask-based defect extraction from registered normal and generated anomaly images.}
  \label{fig:diffmask}
\end{figure}

As detailed in Algorithm~\ref{alg:diffmask}, the registration is refined through similarity alignment, enhanced correlation coefficient (ECC) refinement \cite{ref186},
and, when required, feature-based \cite{ref98} and dense-flow correction \cite{ref187} to absorb
residual geometric drift. The registered images are then compared using
a tolerance band to suppress small registration errors. Five
complementary cues measure luminance, chromaticity, gradient structure,
low-frequency appearance, and fine-detail changes, which are fused into
the difference score $S$. The complete registration and difference-based extraction process is illustrated in~\Cref{fig:diffmask}.

\begin{algorithm}[t]
\caption{DiffMask}
\label{alg:diffmask}
\begin{algorithmic}[1]
\REQUIRE Normal Image $I_n$, Generated Anomaly Image $I_a$
\ENSURE Defect Mask $M_d$, Defect Crop $C_d$

\STATE Convert $I_n$ and $I_a$ to structural representations and obtain an
initial similarity transform using scale--translation search.
\STATE Refine registration using coarse-to-fine ECC alignment; if the
alignment score is insufficient, evaluate feature-based similarity
registration and retain the better solution.
\STATE Apply smoothed dense-flow refinement to compensate for residual
local misalignment.
\STATE Photometrically match the registered $I_n$ to $I_a$ using local
gain--offset fitting and construct a tolerance band around the reference
response.
\STATE Compute luminance, chromaticity,
gradient/edge structure, low-frequency appearance, and sharpness/fine-detail
residuals, and fuse them into the difference score $S$.
\STATE Estimate robust global and local neighborhood statistics of $S$ and
compute the local z-score $Z(p)$.
\STATE Generate strong and weak candidates using absolute and adaptive
thresholds, and apply hysteresis thresholding to retain weak candidates
connected to strong defect evidence.
\STATE Remove isolated responses and refine surviving regions; rank
connected components by accumulated evidence and retain valid ones.
\STATE Recover weak defect continuations using direction-matched completion
and obtain the final defect mask $M_d$.
\STATE Extract the defect crop $C_d$ from $I_a$ with surrounding substrate
context.
\RETURN $M_d$, $C_d$
\end{algorithmic}
\end{algorithm}

Rather than applying a fixed global threshold to $S$, DiffMask evaluates
each pixel relative to its local reconstruction noise. For a pixel $p$,
let $\mathcal{N}(p)$ denote its local neighborhood. The local deviation
is computed as

\begin{equation}
Z(p)=
\frac{S(p)-\mu_{\mathcal{N}(p)}}
{\max\left(\sigma_{\mathcal{N}(p)},\epsilon\right)},
\label{eq:diffmask_zscore}
\end{equation}

where $Z(p)$ is the local z-score, $S(p)$ is the fused difference score,
$\mu_{\mathcal{N}(p)}$ and $\sigma_{\mathcal{N}(p)}$ are the mean and
standard deviation of $S$ within $\mathcal{N}(p)$, and $\epsilon$
prevents unstable normalization when the local variance is near zero.
This makes the defect decision dependent on how strongly a pixel
deviates from its surrounding reconstruction noise rather than on its
absolute difference magnitude.

The resulting $Z$ map is separated into strong and weak candidates using
absolute and adaptive thresholds. Hysteresis thresholding \cite{ref23} retains weak
responses only when connected to strong defect evidence, suppressing
isolated noise while preserving faint defect extensions. The surviving
regions are then refined, ranked by accumulated evidence, and subjected
to direction-matched completion to recover weak portions that exhibit
the same characteristic change as the detected defect. The resulting Defect Mask $M_d$ is used
to extract the Defect Crop $C_d$ from $I_a$ with surrounding substrate
context for subsequent validation and storage.

\subsection{Defect Validation and Storage}
\label{sec:defect_validation}

DiffMask recovers candidate defect regions, but its difference cues can
also flag random texture variations or reconstruction artifacts that do
not correspond to valid defects. Therefore, the extracted Defect Crop
$C_d$ is passed to VLM-2\footnote{Qwen2.5-VL-7B-Instruct \cite{ref9}} for semantic
validation. VLM-2 evaluates $C_d$ together with its surrounding
substrate and accepts it only when the detected region represents a
plausible category-specific defect; otherwise, the crop is rejected. A
contrast test additionally verifies that the candidate defect has
sufficient contrast relative to its surrounding substrate. Each
accepted crop becomes a \textit{Validated Defect} and is stored in the
defect bank $B$, indexed by category and
defect type. Thus, entries are not selected solely based on pixel-level differences; they must also pass VLM-2's semantic validity check and correspond to a valid defect.

Once a Defect Crop $C_d$ is accepted into $B$, its generative cost is
incurred only once. The stored defect can then be retrieved and reused
across different Normal Host Images $I_h$, locations, and spatial
extents, allowing a limited set of generated and validated defects to
produce many distinct synthetic anomaly instances without invoking the
image-generation model again.

\subsection{Object-Aware Localization}
\label{sec:object_aware_localization}

After constructing the defect bank $B$, FLASH must determine where a retrieved defect can be placed on a different Normal Host Image $I_h$. The localization stage consists of two components. First, Object Boundary Suppression (OBS) estimates the physically meaningful foreground region $\Omega$ of $I_h$. Second, Multi-Resolution Spectral Pyramid (MRSP) generates a coherent placement field within this region. The resulting MRSP placement mask $M_f$ controls the location and spatial extent of the retrieved defect without altering its appearance. This allows the same defect to be reused across different host images, locations, and spatial extents.

\subsubsection{Object Boundary Suppression (OBS)}
\label{sec:obs}

The procedural noise field has no knowledge of the physical object in $I_h$ and can therefore place a defect on the background. The key intuition behind OBS is to combine appearance and texture cues: pixels that differ from the background (estimated from the image boundary) are likely to belong to the foreground object \cite{ref1}, while local texture variation can recover object regions that have weak color separation.

As summarized in \Cref{alg:obs}, OBS first estimates the background of each Host Image from the median color of its image-border pixels in CIELAB space. A color-distance response is then Otsu-thresholded \cite{ref108}, while a local standard-deviation response provides an independent texture cue. The two responses are combined using a logical OR and subsequently refined using morphological operations \cite{ref139} and connected-component area filtering. A degeneracy guard handles cases in which the resulting foreground estimate is either empty or covers nearly the complete image.

The resulting Object-Aware region is defined in Eq.~\ref{eq:omega}:

\begin{equation}
\Omega =
\mathcal{R}
\left(
M_{\mathrm{col}}^{\mathrm{Otsu}}
\lor
M_{\mathrm{tex}}^{\mathrm{Otsu}}
\right),
\label{eq:omega}
\end{equation}

where $M_{\mathrm{col}}^{\mathrm{Otsu}}$ and $M_{\mathrm{tex}}^{\mathrm{Otsu}}$ are the color- and texture-based foreground responses, respectively, and $\mathcal{R}(\cdot)$ denotes morphological refinement and connected-component filtering. The OR operation allows a region to be retained when either color or local texture provides sufficient evidence of the object.

\begin{algorithm}[t]
\caption{Object Boundary Suppression}
\label{alg:obs}
\begin{algorithmic}[1]
\REQUIRE Different Normal Host Image $I_h$
\ENSURE Object-Aware region $\Omega$

\STATE Convert $I_h$ to CIELAB space and estimate the background color from the median of the image-border pixels.
\STATE Compute the color-distance response from the estimated background and local grayscale standard-deviation map.
\STATE Apply Otsu thresholding to obtain the color response $M_{\mathrm{col}}^{\mathrm{Otsu}}$ from the estimated background and the texture response $M_{\mathrm{tex}}^{\mathrm{Otsu}}$ from the grayscale standard-deviation map.
\STATE Combine the responses using a logical OR.
\STATE Refine with morphological opening, closing and hole filling.
\STATE Remove connected components below the minimum area threshold; retain the largest component if area filtering removes all candidates.
\STATE Apply the degeneracy guard for empty or near-full-image estimates to obtain $\Omega$ (Eq.~\ref{eq:omega}).
\RETURN $\Omega$
\end{algorithmic}
\end{algorithm}

\subsubsection{Multi-Resolution Spectral Pyramid (MRSP)}
\label{sec:mrsp}

Given the Object-Aware region $\Omega$, FLASH must determine a compact
region within the object where a retrieved defect can be synthesized.
This region should be spatially coherent while allowing both its
location and size to vary across generated samples. MRSP constructs such a placement field by combining two noise components generated across multiple spatial resolutions \cite{ref22}: multi-resolution spectral noise \cite{ref47} and bilinear upsampling noise \cite{ref87}. Low-frequency components establish the coarse blob structure, whereas higher-frequency components introduce irregularity along its boundaries.

As summarized in \Cref{alg:mrsp}, MRSP generates spectral noise at
multiple resolutions using $1/f^\alpha$ amplitude weighting \cite{ref47} and
uniformly sampled random phase. Here, $f$ denotes spatial frequency
and $\alpha$ controls the frequency-dependent amplitude decay. Each
spectral response is transformed to the spatial domain using an
inverse Fourier transform; lower-resolution responses are then
bilinearly upsampled to the host resolution and normalized. The
resulting multi-resolution responses are combined using persistence
weighting \cite{ref114} and energy normalization to obtain the placement field
$F_{\mathrm{MRSP}}$. Thus, spectral synthesis controls the frequency
structure, while bilinear upsampling brings each resolution to a
common spatial scale.

The target extent or size is derived from normal-image statistics. Object coverage, the number of visible object
instances, and the dataset-specific defect-coverage prior determine
the target coverage $c$. Hence, the stochastic MRSP realization
controls \emph{where} the placement appears, while $c$ controls
\emph{how much} of the object it occupies.

A threshold computed over the complete image can select strong
responses outside $\Omega$, while direct intersection with $\Omega$
can fragment or remove the intended placement region. FLASH therefore
computes the threshold only from responses within $\Omega$. Let
$Q_q(\cdot)$ denote the $q$-quantile operator. The final placement mask
is defined in Eq.~\ref{eq:mf}:

\begin{equation}
M_f = \mathcal{L}\!\left[\left(F_{\mathrm{MRSP}} >
Q_{1-c}\!\left(F_{\mathrm{MRSP}}[\Omega>0]\right)\right)
\land\Omega\right],
\label{eq:mf}
\end{equation}

where $F_{\mathrm{MRSP}}$ is the normalized MRSP placement field,
$\Omega$ is the Object-Aware region obtained from OBS,
$Q_{1-c}(\cdot)$ is the $(1-c)$-quantile operator, $c$ is the target
object-region coverage, and $\mathcal{L}(\cdot)$ retains the largest connected
component. Computing the quantile within $\Omega$ makes the specified
coverage relative to the object rather than the complete image.

The resulting $M_f$ provides a single coherent placement region for
Stage~5. Different MRSP realizations vary the defect location, while
different target coverages vary its occupied object area, providing
controllable spatial and size diversity without additional generative
inference.

\begin{algorithm}[t]
\caption{Multi-Resolution Spectral Pyramid}
\label{alg:mrsp}
\begin{algorithmic}[1]
\REQUIRE Host dimensions $(H,W)$, Object-Aware region $\Omega$, target coverage $c$
\ENSURE MRSP placement mask $M_f$

\STATE Construct the multi-resolution frequency grids and generate spectral amplitudes using $1/f^\alpha$ weighting.
\STATE Sample a uniformly random phase for each spectral level and form the spectral response using the inverse Fourier transform.
\STATE Bilinearly resize each response to $(H,W)$ and normalize it.
\STATE Combine the normalized responses using persistence weighting and energy-normalize the combined response to obtain $F_{\mathrm{MRSP}}$.
\STATE Compute the object-conditioned quantile threshold $Q_{1-c}$ from $F_{\mathrm{MRSP}}[\Omega > 0]$.
\STATE Threshold $F_{\mathrm{MRSP}}$, gate by $\Omega$, and retain the largest connected component as $M_f$ (Eq.~\ref{eq:mf}).
\RETURN $M_f$
\end{algorithmic}
\end{algorithm}

\subsection{Adaptive Synthesis}

This final stage combines the Normal Host Image $I_h$, MRSP placement mask $M_f$ and retrieved defect crop $C_d$ to produce the final synthetic anomaly $I_{syn}$. The objective is to reuse the generated defect while varying its placement and size on a new normal host. Thus, $C_d$ determines \emph{what} is synthesized, while $M_f$ determines \emph{where} it is synthesized.

As summarized in Algorithm~\ref{alg:adaptive_synthesis}, FLASH uses an original placement mode and an adaptive placement mode. The original mode places the defect at the MRSP blob centroid while preserving its source-relative scale. The adaptive mode additionally uses the blob area, principal orientation, centroid, coverage, and containment to adapt the defect to the selected MRSP region. The defect is isotropically scaled according to the target coverage, aligned with the MRSP orientation, anchored at the blob centroid or deepest interior point \cite{ref130}, and iteratively reduced when necessary to satisfy the containment constraint. Rotation and horizontal flipping provide additional placement variation while preserving the recovered defect morphology.

\begin{algorithm}[t]
\caption{Adaptive Placement and Hybrid Blending}
\label{alg:adaptive_synthesis}
\begin{algorithmic}[1]
\REQUIRE Host Image $I_h$, Defect Crop $C_d$, Placement Mask $M_f$, Object-Aware region $\Omega$
\ENSURE Synthetic Anomaly $I_{syn}$, Ground-Truth Mask $M_{gt}$

\STATE Extract the MRSP blob centroid $(x_f,y_f)$, area $A_f$, and orientation $\phi_f$.
\STATE Initialize the defect using its source-relative scale and sample rotation and horizontal flip.
\IF{adaptive placement}
    \STATE Estimate defect orientation $\phi_d$ and align with $\phi_f$; isotropically scale the defect according to the MRSP blob coverage.
    \STATE Anchor the defect at $(x_f,y_f)$ or the deepest interior point of $M_f$; reduce its scale until the defect mask satisfies the MRSP containment constraint.
\ELSE
    \STATE Place the defect at $(x_f,y_f)$ using its source-relative scale.
\ENDIF
\STATE Place the transformed defect on $I_h$ and constrain its mask by $M_f \cap \Omega$; harmonize the localized crop with the host substrate in CIELAB space.
\STATE Set $M_{gt}$ to the placed defect mask and compute its equivalent radius $r_{\mathrm{eq}}$.
\IF{$r_{\mathrm{eq}} < \tau$}
    \STATE Apply feathered alpha blending.
\ELSE
    \STATE Extract the defect core and construct a substrate collar around $M_{gt}$ to form $M_c$; apply Poisson blending within $M_c$, falling back to alpha blending if it fails.
\ENDIF
\RETURN $I_{syn}, M_{gt}$
\end{algorithmic}
\end{algorithm}

The placed crop is then harmonized with $I_h$ in CIELAB space using source-substrate and local host-substrate statistics \cite{ref188}. FLASH subsequently applies hybrid blending: small adaptive placements use feathered alpha blending \cite{ref118}, while larger placements use Poisson blending \cite{ref113} with a substrate collar. The collar moves the blending boundary into the surrounding substrate \cite{ref75}, preventing the blending operation from crossing the defect boundary and reducing visible seams that could otherwise be detected as artificial edges. If Poisson blending fails, alpha blending is used as a fallback. The hybrid operation is defined in Eq.~\ref{eq:hybrid_blending}.

\begin{equation}
I_{syn} =
\begin{cases}
\mathcal{A}(I_h,C_d'), & r_{\mathrm{eq}} < \tau,\\
\mathcal{P}(I_h,C_d',M_c), & r_{\mathrm{eq}} \geq \tau,
\end{cases}
\label{eq:hybrid_blending}
\end{equation}

where $I_{syn}$ is the final synthetic anomaly, $I_h$ is the
normal host image, $C_d'$ is the CIELAB-harmonized defect crop,
$r_{\mathrm{eq}}$ is the equivalent radius of the placed defect,
$\tau$ is the small-defect threshold, $\mathcal{A}$ denotes feathered
alpha blending, $\mathcal{P}$ denotes Poisson blending, and $M_c$ is
the substrate-collar compositing mask.

\section{Experiments}
\label{sec:Results_and_Analysis}

This section evaluates synthetic anomalies from FLASH as a substitute for
real defects in decision-threshold calibration across anomaly detectors and
MVTec AD 2 categories, with real-defect calibration as the oracle.

\subsection{Experimental Setup}
\label{sec:experimental_setup}

We evaluate FLASH on MVTec AD 2~\cite{ref60}, which contains high-resolution
industrial inspection scenarios with varying illumination, overlapping
objects, transparent and reflective surfaces, and localized defects. The
complete pipeline is executed independently for each category following the
\emph{Generate Once, Synthesize Many} formulation. SuperADD~\cite{ref190}
with DINOv3 ViT-H/16+~\cite{ref191} features is used as the primary
downstream detector. To determine whether the calibration behavior transfers
beyond a single detector, we additionally evaluate PaDiM~\cite{ref37},
PatchCore~\cite{ref131}, AnomalyDINO~\cite{ref193} and Dinomaly~\cite{dinomaly_cvpr},
covering parametric, coreset-based, training-free, and distillation-based
architectures, respectively, using Anomalib~\cite{ref4}.

For each evaluation, the detector and its memory bank are fitted once and
held fixed; only the data used for decision-threshold calibration is varied.
The \emph{Real} setting fits the threshold using real test anomalies and
therefore serves as an oracle reference rather than a competing method.
Each synthetic setting instead uses held-out normal samples and its own
generated anomalies for calibration and determining the threshold, followed
by evaluation on the same real test set.
A synthetic source is therefore effective
to the extent that its calibrated performance approaches the \emph{Real}
reference. All experiments are repeated with three random seeds and the reported results are averaged across seeds.

\begin{table*}[t]
  \centering

  \begin{minipage}[t]{0.48\textwidth}
    \centering
    \scriptsize
    \resizebox{\linewidth}{!}{%
      \begin{tabular}{@{}lcccc@{}}
        \toprule
        Model & Real & Perlin & AnoStyler & FLASH (Ours) \\
        \midrule
        PaDiM       & 80.37 & 69.41 & 62.37 & \textbf{79.11} \\
        PatchCore   & 82.46 & 67.50 & 43.02 & \textbf{74.64} \\
        AnomalyDINO & 81.47 & 62.19 & 49.85 & \textbf{77.05} \\
 Dinomaly & 81.81& 60.42& 55.37&\textbf{77.51}\\
        SuperADD    & 83.64 & \textbf{79.12} & 64.36 & 78.13 \\
        \bottomrule
      \end{tabular}%
    }
    \vspace{2pt}

    \textbf{(a) Image F1}
  \end{minipage}
  \hfill
  \begin{minipage}[t]{0.48\textwidth}
    \centering
    \scriptsize
    \resizebox{\linewidth}{!}{%
      \begin{tabular}{@{}lcccc@{}}
        \toprule
        Model & Real & Perlin & AnoStyler & FLASH (Ours) \\
        \midrule
        PaDiM       & 7.63  & 3.05  & \textbf{5.37}  & 3.89 \\
        PatchCore   & 26.09 & 14.90 & 16.08 & \textbf{18.14} \\
        AnomalyDINO & 33.55 & 17.47 & 13.99 & \textbf{26.11} \\
        Dinomaly & 31.90& 15.45& 3.35&\textbf{22.53}\\
        SuperADD    & 51.53 & 19.44 & 36.99 & \textbf{38.38} \\
        \bottomrule
      \end{tabular}%
    }
    \vspace{2pt}

    \textbf{(b) Pixel F1}
  \end{minipage}

  \caption{(a) Image-level F1 and (b) pixel-level F1 scores (\%) averaged across categories of the MVTec AD 2.}
  \label{tab:main_results_a}
\end{table*}

\subsection{Cross-Model Comparison}
\label{sec:cross_model}

We compare FLASH with Perlin noise~\cite{ref114,ref170} and
AnoStyler~\cite{ref192} under the same synthetic-anomaly budget and
calibration protocol, with the downstream detector fixed. Perlin represents
the procedural, reference-free baseline, blending textures from the
Describable Textures Dataset through Perlin-noise masks as implemented in
DRAEM~\cite{ref170}.
AnoStyler represents the text-driven generative
alternative, using a style-transfer network under CLIP-guided
textual losses to synthesize defects within procedurally localized
foreground masks. This controlled comparison isolates the effect of the synthetic calibration source on transfer to real defects.

Table~\ref{tab:main_results_a} shows that FLASH provides the most consistent
synthetic calibration across the five detectors. At the image level, FLASH
retains 91--98\% of the corresponding \emph{Real} oracle performance and
achieves the best synthetic result for four of the five detectors. At the pixel
level, FLASH retains 51--78\% of oracle performance and achieves the best
synthetic result for four out of the five detectors. For SuperADD, FLASH reaches 78.1\%
image-level F1 against the 83.6\% \emph{Real} oracle, while substantially
improving pixel-level F1 over Perlin. Although Perlin obtains a marginally
higher image-level F1 for SuperADD, its weaker pixel-level result shows that
global anomaly separation does not necessarily yield spatially useful
calibration. AnoStyler remains stronger for PaDiM at the
pixel level, but exhibits larger variation across detectors. Overall, FLASH
provides the most consistent cross-model transfer rather than dominating
every individual metric.

\begin{table}[t]
    \centering
    \scriptsize
    \setlength{\tabcolsep}{4pt}
    \resizebox{\columnwidth}{!}{%
    \begin{tabular}{@{}lcccc@{}}
        \toprule
        Category & Real & Perlin & AnoStyler & FLASH (Ours) \\
        \midrule
        Can         & 0.02 & \textbf{0.01} & 0.00 & 0.00 \\
        Fabric      & 78.36 & 18.57 & 44.30 & \textbf{69.06} \\
        Fruit Jelly & 56.07 & 40.02 & \textbf{55.93} & 55.57 \\
        Rice        & 58.75 & 6.21 & 25.36 & \textbf{51.90} \\
        Sheet Metal & 35.87 & 2.53 & \textbf{7.79} & 7.39 \\
        Vial        & 57.05 & \textbf{56.46} & 46.63 & 39.41 \\
        Wallplugs   & 54.30 & 1.98 & \textbf{50.79} & 17.41 \\
        Walnuts     & 71.86 & 29.78 & 65.12 & \textbf{66.28} \\
        \midrule
        Mean        & 51.53 & 19.44 & 36.99 & \textbf{38.38}\\
        \bottomrule
    \end{tabular}%
    }
    \caption{Per-category pixel-level F1 scores (\%) using SuperADD. Bold marks the best synthetic source.}
    \label{tab:seg_f1_per_category}
\end{table}

\subsection{Category-wise Behavior and Efficiency}
\label{sec:per_category}

Table~\ref{tab:seg_f1_per_category} examines pixel-level F1 for SuperADD across the
eight MVTec AD 2 categories, with \emph{Real} providing the oracle reference.
FLASH achieves the highest mean among the synthetic sources and approaches
the oracle most closely on several textured-material categories, including
fabric, walnuts, rice, and fruit jelly. The improvement is particularly
pronounced for rice, where Perlin provides substantially weaker calibration.
The remaining categories reveal complementary behavior: AnoStyler performs
better on wallplugs, where the anomalies are primarily geometric, while
Perlin performs better on vial, a transparent back-lit setting. Sheet metal
remains challenging for all synthetic sources under its dark-field
illumination and specular structure. The \emph{can} category is a hard case,
with the \emph{Real} oracle itself achieving close to 0, and therefore does not
provide a meaningful distinction between synthesis strategies. Failure on this category is reported by SuperADD in \cite{ref190} as well.
These results support FLASH as a transferable calibration source across
defect types, while identifying where alternative synthesis strategies
remain advantageous.

Table~\ref{tab:image_gen_times} shows that Perlin and AnoStyler require no
separate $I_a$ generation, whereas FLASH incurs this one-time step for defect
extraction. Assuming three donor images ($I_a$) at 15~s each and 90 synthesis replays,
FLASH has an estimated generation time of 97.8~s, compared with 1348.4~s
for AnoStyler. This gives an estimated 13.8$\times$ speedup or a 92.8\%
reduction in generation time.

\begin{table}[t]
    \centering
    \scriptsize
    \setlength{\tabcolsep}{3.5pt}
    \resizebox{\columnwidth}{!}{%
    \begin{tabular}{@{}lccc@{}}
        \toprule
        Method & $I_a$ Gen. (s/image) & $I_{syn}$ Gen. (ms/image) & Total (s/category) \\
        \midrule
        Perlin       & --  & 14.4    & 1.3 \\
        AnoStyler    & --  & 14,982.1 & 1,348.4 \\
        FLASH (Ours) & 15  & 586.2   & 97.8 \\
        \bottomrule
    \end{tabular}%
    }
     \caption{Estimated generation cost for 90 samples at native resolutions: Perlin $448^2$, AnoStyler $256^2$, and FLASH $1024^2$. FLASH assumes three donor images $I_a$ at 15~s each; totals exclude model loading, VLM checks, DiffMask recovery, and bank extraction. Ratios use unrounded totals.}
    \label{tab:image_gen_times}
\end{table}

\section{Conclusion and Future Work}
\label{sec:conclusion}

The results establish FLASH as an effective approach to bridging the
procedural--generative trade-off in synthetic anomaly generation. Across
MVTec AD 2, FLASH provides strong and consistent calibration across diverse
anomaly-detection architectures and categories, reaching 78.1\% image-level
F1 for SuperADD against an 83.6\% Real oracle reference. Its reusable defect
banks preserve the contextual diversity of generative anomalies while
enabling their transfer across normal hosts, making synthetic anomalies a
viable surrogate for real defects in decision-threshold calibration. FLASH
also provides an estimated 13.8$\times$ image-generation speedup over
AnoStyler (a 92.8\% reduction), under the timing assumptions and exclusions
in Table~\ref{tab:image_gen_times}.
Together, these findings establish that FLASH can provide scalable synthetic
calibration while substantially reducing dependence on real defect samples.

Future work will focus on reducing the remaining category-dependent
calibration gaps, particularly for geometry-dominated and challenging
illumination regimes, through increased defect diversity and adaptive
defect-bank construction. We will further evaluate FLASH across additional
industrial datasets and study how calibration quality evolves with larger
synthetic anomaly sets.
{
    \small
    \bibliographystyle{ieeenat_fullname}
    \bibliography{main}

@string{CVPR   = {CVPR}}

@string{ICCV   = {ICCV}}

@string{ECCV   = {ECCV}}

@string{AAAI   = {AAAI}}

@string{ICLR   = {ICLR}}

@string{ICPR   = {ICPR}}

@string{ICIP   = {ICIP}}

@string{PAMI   = {IEEE Trans. Pattern Anal. Mach. Intell.}}

@string{IJCV   = {Int. J. Comput. Vis.}}

@string{TOG    = {ACM Trans. Graph.}}

@inproceedings{ref1,
  author    = {R. Achanta and S. Hemami and F. Estrada and S. Süsstrunk},
  title     = {Frequency-tuned salient region detection},
  booktitle = CVPR,
  pages     = {1597--1604},
  year      = {2009}
}

@inproceedings{ref4,
  author    = {S. Akcay and D. Ameln and A. Vaidya and B. Lakshmanan and N. Ahuja and U. Genc},
  title     = {Anomalib: A deep learning library for anomaly detection},
  booktitle = ICIP,
  pages     = {1706--1710},
  year      = {2022}
}

@article{ref9,
  author    = {S. Bai and K. Chen and X. Liu and others},
  title     = {Qwen2.5-VL technical report},
  journal   = {arXiv:2502.13923},
  year      = {2025}
}

@inproceedings{ref10,
  author    = {K. Batzner and L. Heckler and R. König},
  title     = {EfficientAD: Accurate visual anomaly detection at millisecond-level latencies},
  booktitle = {WACV},
  pages     = {128--138},
  year      = {2024}
}

@inproceedings{ref11,
  author    = {P. Bergmann and M. Fauser and D. Sattlegger and C. Steger},
  title     = {MVTec AD — A comprehensive real-world dataset for unsupervised anomaly detection},
  booktitle = CVPR,
  pages     = {9592--9600},
  year      = {2019}
}

@article{ref22,
  author    = {P. J. Burt and E. H. Adelson},
  title     = {The Laplacian pyramid as a compact image code},
  journal   = {IEEE Trans. Communications},
  volume    = {31},
  number    = {4},
  pages     = {532--540},
  year      = {1983}
}

@article{ref23,
  author    = {J. Canny},
  title     = {A computational approach to edge detection},
  journal   = PAMI,
  volume    = {PAMI-8},
  number    = {6},
  pages     = {679--698},
  year      = {1986}
}

@inproceedings{ref26,
  author    = {Q. Chen and H. Luo and C. Lv and Z. Zhang},
  title     = {A unified anomaly synthesis strategy with gradient ascent for industrial anomaly detection and localization},
  booktitle = ECCV,
  pages     = {37--54},
  year      = {2024}
}

@inproceedings{ref36,
  author    = {Z. Dai and S. Zeng and H. Liu and X. Li and F. Xue and Y. Zhou},
  title     = {SeaS: Few-shot industrial anomaly image generation with separation and sharing fine-tuning},
  booktitle = ICCV,
  year      = {2025}
}

@inproceedings{ref37,
  author    = {T. Defard and A. Setkov and A. Loesch and R. Audigier},
  title     = {PaDiM: A patch distribution modeling framework for anomaly detection and localization},
  booktitle = {ICPR Workshops},
  pages     = {475--489},
  year      = {2021}
}

@inproceedings{ref38,
  author    = {H. Deng and X. Li},
  title     = {Anomaly detection via reverse distillation from one-class embedding},
  booktitle = CVPR,
  pages     = {9737--9746},
  year      = {2022}
}

@inproceedings{ref40,
  author    = {Y. Duan and Y. Hong and L. Niu and L. Zhang},
  title     = {Few-shot defect image generation via defect-aware feature manipulation},
  booktitle = AAAI,
  pages     = {571--578},
  year      = {2023}
}

@article{ref47,
  author    = {A. Fournier and D. Fussell and L. Carpenter},
  title     = {Computer rendering of stochastic models},
  journal   = {Communications of the ACM},
  volume    = {25},
  number    = {6},
  pages     = {371--384},
  year      = {1982}
}

@inproceedings{ref53,
  author    = {Z. Gu and B. Zhu and G. Zhu and Y. Chen and M. Tang and J. Wang},
  title     = {AnomalyGPT: Detecting industrial anomalies using large vision-language models},
  booktitle = AAAI,
  pages     = {1932--1940},
  year      = {2024}
}

@inproceedings{ref54,
  author    = {G. Gui and B.-B. Gao and J. Liu and C. Wang and Y. Wu},
  title     = {Few-shot anomaly-driven generation for anomaly classification and segmentation},
  booktitle = ECCV,
  pages     = {210--226},
  year      = {2024}
}

@article{ref60,
  author    = {L. Heckler-Kram and J.-H. Neudeck and U. Scheler and R. König and C. Steger},
  title     = {The MVTec AD 2 dataset: Advanced scenarios for unsupervised anomaly detection},
  journal   = {arXiv:2503.21622},
  year      = {2025}
}

@article{ref66,
  author    = {J. Hu and F. Borsatti and A. Stropeni and D. Dalle Pezze and M. Barusco and G. A. Susto},
  title     = {MIRAGE: Model-agnostic industrial realistic anomaly generation and evaluation},
  journal   = {arXiv:2603.13507},
  year      = {2026}
}

@inproceedings{ref67,
  author    = {T. Hu and others},
  title     = {AnomalyDiffusion: Few-shot anomaly image generation with diffusion model},
  booktitle = AAAI,
  pages     = {8526--8534},
  year      = {2024}
}

@inproceedings{ref72,
  author    = {J. Jeong and Y. Zou and T. Kim and D. Zhang and A. Ravichandran and O. Dabeer},
  title     = {WinCLIP: Zero-/few-shot anomaly classification and segmentation},
  booktitle = CVPR,
  pages     = {19606--19616},
  year      = {2023}
}

@article{ref75,
  author    = {J. Jia and J. Sun and C.-K. Tang and H.-Y. Shum},
  title     = {Drag-and-drop pasting},
  journal   = TOG,
  volume    = {25},
  number    = {3},
  pages     = {631--637},
  year      = {2006}
}

@inproceedings{ref77,
  author    = {Y. Jiang and others},
  title     = {Anomagic: Crossmodal prompt-driven zero-shot anomaly generation},
  booktitle = AAAI,
  pages     = {5485--5493},
  year      = {2026}
}

@inproceedings{ref78,
  author    = {Y. Jin and others},
  title     = {Dual-interrelated diffusion model for few-shot anomaly image generation},
  booktitle = CVPR,
  year      = {2025}
}

@inproceedings{ref86,
  author    = {Z. Lai and others},
  title     = {AnomalyPainter: Vision-language-diffusion synergy for realistic and diverse unseen industrial anomaly synthesis},
  booktitle = AAAI,
  pages     = {5800--5808},
  year      = {2026}
}

@article{ref87,
  author    = {J. P. Lewis},
  title     = {Algorithms for solid noise synthesis},
  journal   = {ACM SIGGRAPH Computer Graphics},
  volume    = {23},
  number    = {3},
  pages     = {263--270},
  year      = {1989}
}

@inproceedings{ref89,
  author    = {C.-L. Li and K. Sohn and J. Yoon and T. Pfister},
  title     = {CutPaste: Self-supervised learning for anomaly detection and localization},
  booktitle = CVPR,
  pages     = {9664--9674},
  year      = {2021}
}

@inproceedings{ref97,
  author    = {Z. Liu and Y. Zhou and Y. Xu and Z. Wang},
  title     = {SimpleNet: A simple network for image anomaly detection and localization},
  booktitle = CVPR,
  pages     = {20402--20411},
  year      = {2023}
}

@article{ref98,
  author    = {D. G. Lowe},
  title     = {Distinctive image features from scale-invariant keypoints},
  journal   = IJCV,
  volume    = {60},
  number    = {2},
  pages     = {91--110},
  year      = {2004}
}

@article{ref106,
  author    = {S. Niu and B. Li and X. Wang and H. Lin},
  title     = {Defect image sample generation with GAN for improving defect recognition},
  journal   = {IEEE Trans. Automation Science and Engineering},
  volume    = {17},
  number    = {3},
  pages     = {1611--1622},
  year      = {2020}
}

@article{ref108,
  author    = {N. Otsu},
  title     = {A threshold selection method from gray-level histograms},
  journal   = {IEEE Trans. Systems, Man, and Cybernetics},
  volume    = {9},
  number    = {1},
  pages     = {62--66},
  year      = {1979}
}

@article{ref113,
  author    = {P. Pérez and M. Gangnet and A. Blake},
  title     = {Poisson image editing},
  journal   = TOG,
  volume    = {22},
  number    = {3},
  pages     = {313--318},
  year      = {2003}
}

@inproceedings{ref114,
  author    = {K. Perlin},
  title     = {An image synthesizer},
  booktitle = {SIGGRAPH},
  pages     = {287--296},
  year      = {1985}
}

@article{ref118,
  author    = {T. Porter and T. Duff},
  title     = {Compositing digital images},
  journal   = {ACM SIGGRAPH Computer Graphics},
  volume    = {18},
  number    = {3},
  pages     = {253--259},
  year      = {1984}
}

@inproceedings{ref129,
  author    = {R. Rombach and A. Blattmann and D. Lorenz and P. Esser and B. Ommer},
  title     = {High-resolution image synthesis with latent diffusion models},
  booktitle = CVPR,
  pages     = {10684--10695},
  year      = {2022}
}

@article{ref130,
  author    = {A. Rosenfeld and J. L. Pfaltz},
  title     = {Sequential operations in digital picture processing},
  journal   = {Journal of the ACM},
  volume    = {13},
  number    = {4},
  pages     = {471--494},
  year      = {1966}
}

@inproceedings{ref131,
  author    = {K. Roth and L. Pemula and J. Zepeda and B. Schölkopf and T. Brox and P. Gehler},
  title     = {Towards total recall in industrial anomaly detection},
  booktitle = CVPR,
  pages     = {14318--14328},
  year      = {2022}
}

@inproceedings{ref138,
  author    = {H. M. Schlüter and J. Tan and B. Hou and B. Kainz},
  title     = {Natural synthetic anomalies for self-supervised anomaly detection and localization},
  booktitle = ECCV,
  pages     = {474--489},
  year      = {2022}
}

@book{ref139,
  author    = {J. Serra},
  title     = {Image Analysis and Mathematical Morphology},
  publisher = {Academic Press},
  year      = {1982}
}

@inproceedings{ref140,
  author    = {Q. Shi and J. Wei and F. Shen and Z. Zhang},
  title     = {Few-shot defect image generation based on consistency modeling},
  booktitle = ECCV,
  year      = {2024}
}

@inproceedings{ref146,
  author    = {J. Song and others},
  title     = {DefectFill: Realistic defect generation with inpainting diffusion model for visual inspection},
  booktitle = CVPR,
  pages     = {18718--18727},
  year      = {2025}
}

@article{ref152,
  author    = {J. Tan and B. Hou and J. Batten and H. Qiu and B. Kainz},
  title     = {Detecting outliers with foreign patch interpolation},
  journal   = {Machine Learning for Biomedical Imaging},
  volume    = {1},
  pages     = {1--27},
  year      = {2022}
}

@article{ref165,
  author    = {X. Xu and others},
  title     = {A survey on industrial anomalies synthesis},
  journal   = {arXiv:2502.16412},
  year      = {2025}
}

@article{ref166,
  author    = {M. Yang and P. Wu and H. Feng},
  title     = {MemSeg: A semi-supervised method for image surface defect detection using differences and commonalities},
  journal   = {Engineering Applications of Artificial Intelligence},
  volume    = {119},
  pages     = {105835},
  year      = {2023}
}

@inproceedings{ref170,
  author    = {V. Zavrtanik and M. Kristan and D. Skočaj},
  title     = {DRAEM — A discriminatively trained reconstruction embedding for surface anomaly detection},
  booktitle = ICCV,
  pages     = {8330--8339},
  year      = {2021}
}

@inproceedings{ref171,
  author    = {G. Zhang and K. Cui and T.-Y. Hung and S. Lu},
  title     = {Defect-GAN: High-fidelity defect synthesis for automated defect inspection},
  booktitle = {WACV},
  pages     = {2524--2534},
  year      = {2021}
}

@article{ref174,
  author    = {Q. Zhang and S. Zhang and J. Liu and others},
  title     = {ASBench: Image anomalies synthesis benchmark for anomaly detection},
  journal   = {arXiv:2510.07927},
  year      = {2025}
}

@inproceedings{ref177,
  author    = {X. Zhang and M. Xu and X. Zhou},
  title     = {RealNet: A feature selection network with realistic synthetic anomaly for anomaly detection},
  booktitle = CVPR,
  pages     = {16699--16708},
  year      = {2024}
}

@inproceedings{ref182,
  author    = {Q. Zhou and G. Pang and Y. Tian and S. He and J. Chen},
  title     = {AnomalyCLIP: Object-agnostic prompt learning for zero-shot anomaly detection},
  booktitle = ICLR,
  year      = {2024}
}

@inproceedings{ref184,
  author    = {Y. Zou and J. Jeong and L. Pemula and D. Zhang and O. Dabeer},
  title     = {SPot-the-difference self-supervised pre-training for anomaly detection and segmentation},
  booktitle = ECCV,
  pages     = {392--408},
  year      = {2022}
}

@article{ref185,
    author  = {OpenAI},
    title   = {{GPT-4o} System Card},
    journal = {arXiv:2410.21276},
    year    = {2024}
  }

@article{ref186,
    author  = {Georgios D. Evangelidis and Emmanouil Z. Psarakis},
    title   = {Parametric Image Alignment Using Enhanced Correlation
               Coefficient Maximization},
    journal = PAMI,
    volume  = {30},
    number  = {10},
    pages   = {1858--1865},
    year    = {2008}
  }

@inproceedings{ref187,
    author    = {Gunnar Farneb{\"a}ck},
    title     = {Two-Frame Motion Estimation Based on Polynomial Expansion},
    booktitle = {SCIA},
    pages     = {363--370},
    year      = {2003}
  }

@article{ref188,
    author  = {Erik Reinhard and Michael Ashikhmin and Bruce Gooch and
               Peter Shirley},
    title   = {Color Transfer Between Images},
    journal = {IEEE Computer Graphics and Applications},
    volume  = {21},
    number  = {5},
    pages   = {34--41},
    year    = {2001}
  }

@inproceedings{ref190,
    author    = {Lukas Roming and Felix Lehnerer and Jonas V. Funk and
                 Andreas Michel and Georg Maier and Thomas L{\"a}ngle and
                 J{\"u}rgen Beyerer},
    title     = {{SuperADD}: Training-Free Class-Agnostic Anomaly Segmentation},
    booktitle = {CVPR Workshops (VAND 4.0 Challenge, Industrial Track)},
    year      = {2026}
  }

@article{ref191,
  author  = {Oriane Sim\'{e}oni and Huy V. Vo and Maximilian Seitzer and
             Federico Baldassarre and Maxime Oquab and others},
  title   = {{DINOv3}},
  journal = {arXiv:2508.10104},
  year    = {2025}
}

@inproceedings{ref192,
  author    = {Yulim So and Seokho Kang},
  title     = {{AnoStyler}: Text-Driven Localized Anomaly Generation via Lightweight Style Transfer},
  booktitle = AAAI,
  year      = {2026},
  note      = {arXiv:2511.06687}
}

@inproceedings{ref193,
  author    = {Simon Damm and Mike Laszkiewicz and Johannes Lederer and Asja Fischer},
  title     = {{AnomalyDINO}: Boosting Patch-based Few-shot Anomaly Detection with {DINOv2}},
  booktitle = {WACV},
  pages     = {1319--1329},
  year      = {2025}
}

@inproceedings{dinomaly_cvpr,
  title={Dinomaly: The less is more philosophy in multi-class unsupervised anomaly detection},
  author={Guo, Jia and Lu, Shuai and Zhang, Weihang and Chen, Fang and Li, Huiqi and Liao, Hongen},
  booktitle=CVPR,
  pages={20405--20415},
  year={2025}
}
}
\clearpage
\section*{Supplementary Material}
\label{sec:supplementary}

This supplementary material describes implementation details, design choices, and additional experiments for FLASH. It covers semantic defect generation, object-aware placement, MRSP, threshold calibration, and defect reuse across categories. 

\subsection*{A. Implementation Details and Hyperparameters}
\label{sec:supp_implementation}

FLASH uses a fixed, category-agnostic configuration across the evaluated categories. The complete implementation is provided with the supplementary code, while Table~\ref{tab:flash_core_params} reports the principal hyperparameters that directly control defect extraction, object-aware placement, adaptive synthesis, and hybrid blending. The configuration was established during development using controlled parameter sweeps and observed failure modes, and was subsequently frozen rather than tuned independently for each category.

\begin{table*}[t]
  \centering
  \small
  \setlength{\tabcolsep}{6pt}
  \renewcommand{\arraystretch}{1.08}
  \begin{tabular}{@{}p{0.29\textwidth}p{0.17\textwidth}p{0.42\textwidth}@{}}
    \toprule
    \textbf{Hyperparameter} & \textbf{Value} & \textbf{Purpose} \\
    \midrule
    Synthesis resolution & $1024$ px & Provides sufficient spatial support for localized defects and blending. \\
    DiffMask context expansion & $1.8$ & Retains surrounding substrate context around the recovered defect. \\
    Minimum defect area & $500$ pixels & Suppresses small reconstruction artifacts during defect extraction. \\
    Defect contrast threshold & $\Delta E=6.0$ & Filters weak defect evidence using perceptual color difference. \\
    MRSP noise scale & $14$ & Controls the characteristic spatial scale of the placement field. \\
    MRSP pyramid levels $L$ & $6$ & Provides multi-resolution spatial structure for defect placement. \\
    MRSP persistence $\rho$ & $0.8$ & Controls the contribution of successive spectral resolutions. \\
    Target object coverage $c$ & $0.018$ & Controls placement extent relative to the detected object. \\
    Adaptive placement fraction & $0.5$ & Balances original and adaptive placement across generated samples. \\
    Defect-area factor $\gamma$ & $\mathcal{U}(0.20,0.60)$ & Controls defect extent relative to the selected placement region. \\
    Containment threshold & $0.90$ & Keeps the transformed defect predominantly within the valid placement region. \\
    Poisson collar fraction & $0.25$ & Provides local substrate support for Poisson blending. \\
    Minimum Poisson collar & $6$ px & Ensures sufficient blending support for small defects. \\
    Poisson routing threshold & $16$ px & Routes insufficiently supported defects to alpha blending. \\
    \bottomrule
  \end{tabular}
  \caption{Principal hyperparameters governing defect extraction, object-aware placement, adaptive defect sizing, and hybrid blending. The configuration is fixed across categories after development.}
  \label{tab:flash_core_params}
\end{table*}

\begin{table}[t]
\centering
\scriptsize
\begin{tabular}{ll}
\toprule
\textbf{Component} & \textbf{Configuration} \\
\midrule
Anomalib & 2.5.2 \\
Python & 3.13 \\
PyTorch & 2.13.0 (CUDA 13.0) \\
PyTorch Lightning & 2.6.5 \\
TorchMetrics & 1.9.0 \\
Torchvision & 0.28.0 \\
timm & 1.0.28 \\
GPU & NVIDIA RTX 3090 (24 GB) \\
CPU & Intel Core i9-10920X @ 3.50 GHz \\
OS & Ubuntu 24.04 LTS \\
\bottomrule
\end{tabular}
\caption{Software and hardware environment used for the main calibration experiments.}
\label{tab:supp-env}
\end{table}

The parameterization follows the intended operating behavior of each stage. DiffMask is designed to retain coherent changes introduced by the generator while suppressing reconstruction noise; consequently, defect size, contextual support, and perceptual contrast are controlled jointly. For placement, MRSP provides multi-scale spatial variation, while its coverage is defined relative to the detected object rather than the image frame. Adaptive placement then varies defect extent within a bounded relative range and enforces geometric containment. For compositing, the Poisson collar provides the substrate support required for seamless blending, while the routing threshold avoids applying Poisson blending when the defect does not provide sufficient spatial support. The principal parameters therefore depend on local image statistics, object geometry, or relative defect scale rather than category-specific coordinates.

The configuration was not subsequently retuned for individual categories. It was fixed using four development categories, \texttt{rice}, \texttt{walnuts}, \texttt{wallplugs}, and \texttt{fruit\_jelly}, and then applied unchanged to four held-out categories, \texttt{can}, \texttt{fabric}, \texttt{sheet\_metal}, and \texttt{vial}. What is held out is the numeric parameterization: no value in Table~\ref{tab:flash_core_params} was retuned for the second group, although the Config File of Section~B supplies category vocabulary for all eight categories. Thus, the held-out categories are processed using the same generation parameters despite differences in object geometry, surface texture, and appearance. This protocol tests whether the fixed parameterization transfers to categories outside the development set.

\begin{figure*}[t]
  \centering
  \includegraphics[width=\textwidth]{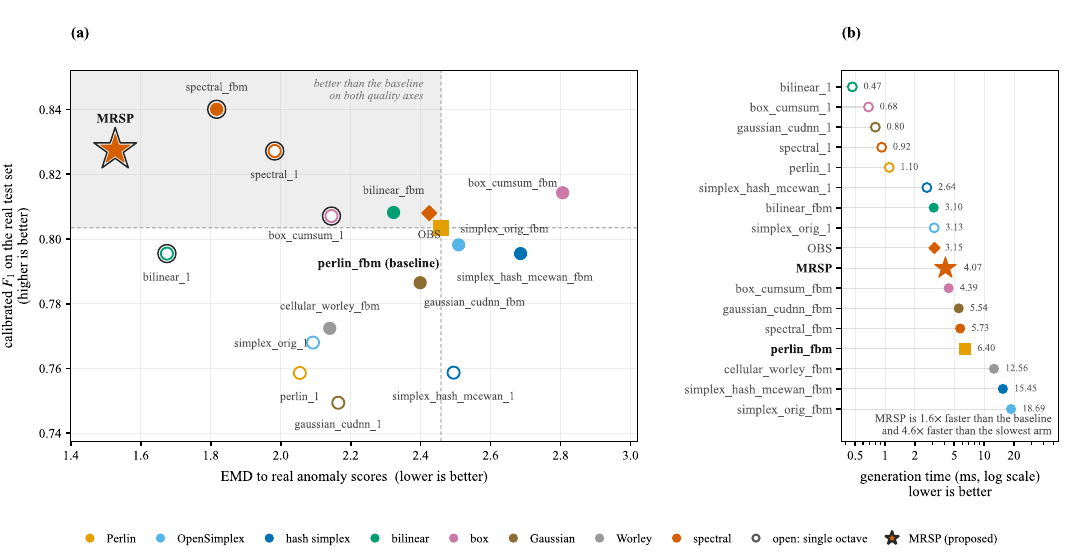}
  \caption{Joint comparison of noise primitives in terms of generation time, earth mover's distance (EMD), and calibrated $F_1$. (a) Calibrated $F_1$ on the real test set versus EMD to the real anomaly-score distribution; lower EMD and higher $F_1$ are preferred. (b) Noise-generation time on a logarithmic scale (measured on images at $224\times224$ resolution), where lower values are preferred. Marker styles distinguish single-octave and six-octave fractal Brownian motion (fBm) configurations, while MRSP is highlighted as the proposed primitive. Here, OBS denotes the band-orthogonal spectral control, not Object Boundary Suppression. MRSP provides a favorable trade-off among computational cost, similarity to real anomaly-score distributions, and calibration transfer.}

  \label{fig:noise_pareto}
\end{figure*}

The same design principle motivates the expected applicability of FLASH to other industrial anomaly datasets with similar structured objects or surfaces and localized defects. Since the principal controls are expressed through local statistics, normalized multi-scale structure, object-relative coverage, defect-relative extent, and geometric containment, they do not depend on manually specified defect locations or category-specific coordinates. Cross-dataset validation is not part of the present study and is reserved for future work.

For the downstream experiments, the corpus and detector configuration are also fixed as summarized in Table~\ref{tab:evaluation_config}. FLASH generates $90$ synthetic anomalies per category and generation seed, except $60$ for fruit jelly and $105$ for vial, matching the public-test anomaly counts. Generation seeds $\{0,1,2\}$ correspond to evaluation seed labels $\{1,2,3\}$. SuperADD uses the frozen DINOv3 ViT-H/16+ backbone and a fixed normal memory bank. These settings remain identical across evaluation settings, so the comparison changes the threshold-fitting data without changing the underlying detector.

The main experiment results for FLASH were generated and evaluated on the dedicated workstation configuration outlined in Table~\ref{tab:supp-env}.

\begin{table*}[t]
  \centering
  \small
  \setlength{\tabcolsep}{7pt}
  \renewcommand{\arraystretch}{1.08}
  \begin{tabular}{@{}p{0.30\textwidth}p{0.18\textwidth}p{0.42\textwidth}@{}}
    \toprule
    \textbf{Configuration} & \textbf{Value} & \textbf{Purpose} \\
    \midrule
    Generation seeds & $\{0,1,2\}$ & Provides repeated paired evaluations under controlled stochastic variation. \\
    VLM & Qwen2.5-VL-7B-Instruct & Used for semantic anomaly specification and defect validation. \\
    Synthetic anomalies & $60$--$105$ / category / seed & Fixed category-specific calibration budgets. \\
    SuperADD backbone & DINOv3 ViT-H/16+ & Frozen representation used across all evaluation settings. \\
    Patch-feature cap & $8000$ & Fixed memory-bank capacity across evaluation settings. \\
    \bottomrule
  \end{tabular}
  \caption{Fixed corpus and downstream evaluation configuration used in the reported experiments.}
  \label{tab:evaluation_config}
\end{table*}

\subsection*{B. Semantic Specification of the Anomaly Prompt}
\label{sec:supp_prompt}

Stage~1 converts a defect-free image $I_n$ into an \emph{Anomaly Prompt} describing one physically plausible defect. Its behavior is controlled by two artifacts: a dataset-specific Config File and a frozen Prompt Template. The Config File specifies the dataset context and, for each category, the object identity, permitted defect families, concrete defect candidates, and category-specific notes describing normal visual variation. An optional \texttt{max\_pixels} override handles extreme aspect ratios. A shared defect-family dictionary and validation vocabulary remain dataset-agnostic, making adaptation to a new dataset primarily a configuration change. Defect size is specified qualitatively (e.g., \texttt{tiny} or \texttt{small}) rather than using physical units.

VLM-1 returns four fields: \texttt{defect\_name}, \texttt{target}, \texttt{defect}, and \texttt{where}, which are inserted into a frozen template:

\begin{quote}
\ttfamily\scriptsize
Generate a natural-looking anomaly image of this scene: \{target\}
has a \{size\} \{defect\}, \{where\}. The defect is part of the
material, not laid on top -- its edges, depth and shadow follow the
surface. Small, clearly visible, photorealistic.
\end{quote}

The template was chosen to avoid two observed failure modes: preservation-dominated prompts that reproduce the input unchanged, and copy-and-overlay prompts that place an external object over the image rather than modifying the material. Three fixed rules are appended to every prompt: (i) preserve geometry, composition, background, texture, lighting, shadows, reflections, perspective, and pixel correspondence outside the defect; (ii) preserve photometric properties, dimensions, aspect ratio, crop, and resolution without global modification; and (iii) introduce exactly one small, category-valid, physically realistic defect. The first two encourage comparability between $I_n$ and the generated anomaly $I_a$, while the third supports single-defect mask recovery.

Each VLM response is validated for schema completeness, category validity, object identity, and defect specificity before generation. Responses are rejected when they describe normal variation, use ambiguous appearance terms, fail the required noun-phrase structure, or reuse a \texttt{defect\_name} within a category. Rejected responses are retried with the rejection reason appended; the first attempt is greedy and subsequent attempts use temperature $0.8$, with four attempts allowed. A deterministic Config File fallback is used if all attempts fail, while the raw response is retained for diagnosis. Three normal images per category serve as generation donors and are assigned different permitted defect families. Each prompt, seed, family, image identifier, and raw response is recorded in \texttt{prompts.csv} and \texttt{manifest.json} for reproducibility.

\subsection*{C. Construction of the Placement Mask $M_f$}
\label{sec:supp_placement_mask}

\paragraph{Noise primitive selection}
The placement mask $M_f$ determines where and over what extent a retrieved defect is placed on the host object, but not the defect appearance. We therefore treat the underlying stochastic field as a design variable and benchmark 17 GPU-based configurations spanning Perlin, OpenSimplex, hash-simplex, bilinear, box, Gaussian, Worley, and spectral fields at single-octave and six-octave fBm depths, together with MRSP and a band-orthogonal spectral control. All candidates use the same mask-construction pipeline and matched coverage, and are evaluated by generation time, the EMD between synthetic and real anomaly scores, and transfer of an $F_1$ threshold calibrated on synthetic anomalies to real test anomalies. The comparison reveals a trade-off between speed and similarity of anomaly-score distributions: bilinear fields are the fastest, spectral fields provide closer score distributions, while multi-octave fBm incurs substantially higher generation cost. We therefore seek a multi-scale field that retains the benefits of spectral structure without repeated spectral evaluation.

MRSP provides this operating point. As shown in Figure~\ref{fig:noise_pareto}, MRSP requires $4.07$\,ms per field versus $6.40$\,ms for six-octave Perlin fBm, while achieving an EMD of $1.53$ and calibrated $F_1$ of $0.828$, compared with $2.46$ and $0.803$, respectively, for the baseline. Thus, relative to Perlin fBm, MRSP improves runtime, score-distribution similarity, and calibration transfer simultaneously, while achieving the lowest EMD among the 17 configurations. Five configurations are non-dominated in the joint comparison, with MRSP among them. We therefore adopt MRSP as the placement primitive for subsequent experiments and use six-octave Perlin fBm as the primary procedural baseline because its multi-scale structure provides a stronger comparison than single-octave Perlin.

\begin{figure}[t]
  \centering
  \includegraphics[width=\columnwidth]{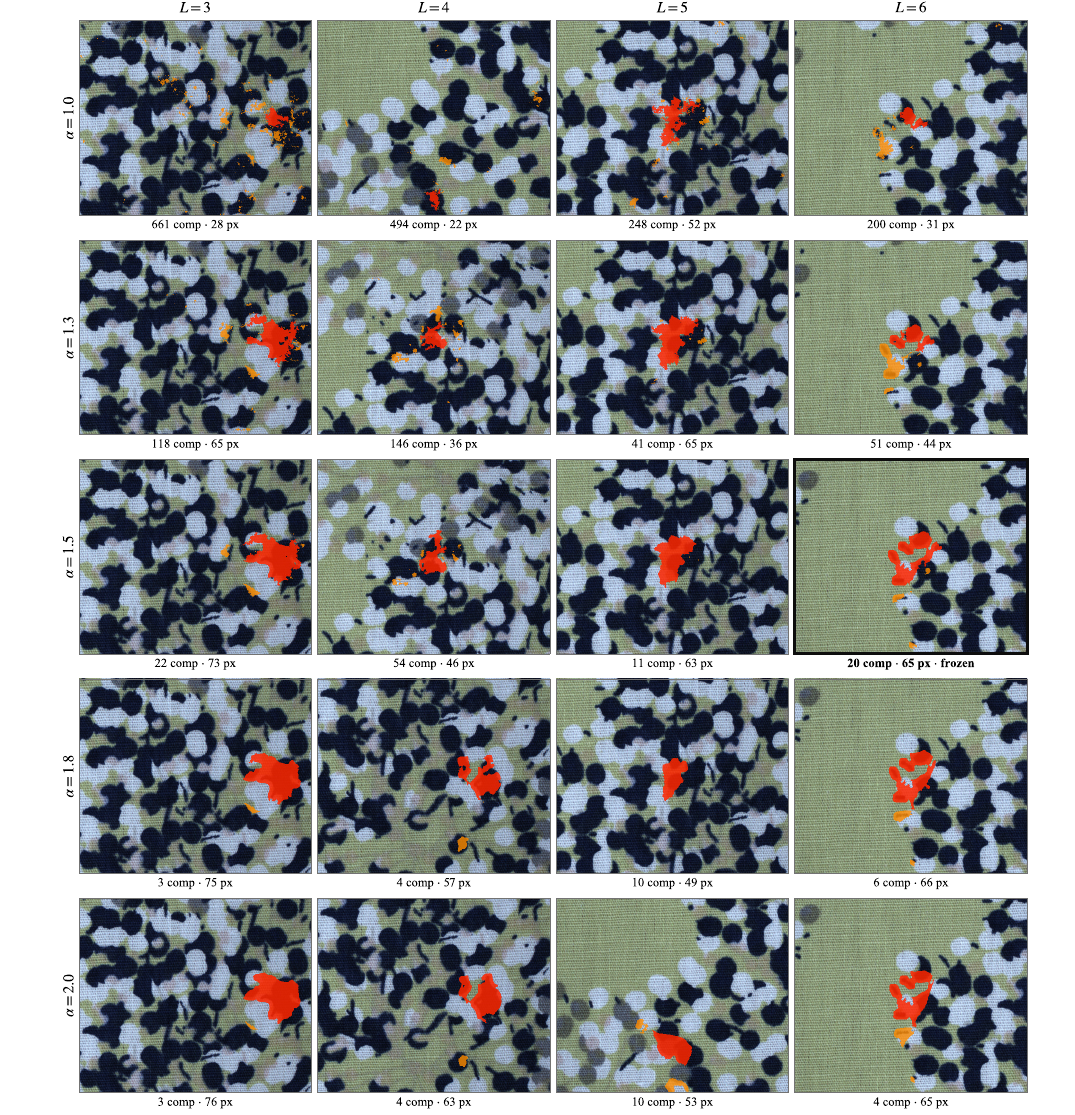}
  \caption{MRSP ablation at the MVTec AD 2 operating point. Placement masks are evaluated across spectral exponent $\alpha$ and pyramid levels $L$ at fixed coverage $c=0.018$. The selected configuration, $\alpha=1.5$ and $L=6$, provides a coherent placement region while avoiding excessive fragmentation or over-smoothing.}
  \label{fig:mrsp_ablation}
\end{figure}

\paragraph{MRSP ablation}
MRSP exposes two parameters: the spectral exponent $\alpha$, which controls the decay of amplitude with frequency and hence the amount of fine-scale structure, and the number of pyramid levels $L$, which determines the available coarse-scale structure. Figure~\ref{fig:mrsp_ablation} evaluates these parameters at the MVTec AD 2 operating point, using Fabric as a representative category and fixing the coverage at $c=0.018$ in every cell. Thus, $\alpha$ and $L$ determine the morphology of the placement region, while coverage independently controls its extent. This low-coverage setting is evaluated directly rather than inherited from the earlier $12\%$ study done on MVTec AD, since thresholding within the object region $\Omega$ produces a different fragmentation regime. Low $\alpha$ values retain strong high-frequency variation and produce fragmented regions, whereas high $\alpha$ values excessively smooth the field into compact, near-circular regions. The effect of $L$ on fragmentation is non-monotonic; we fix $L=6$ for subsequent experiments.

The choice of $\alpha$ is determined by the transition from fragmentation to excessive smoothness. At $\alpha=1.5$ and $L=6$, fragmentation has largely collapsed, with 20 thresholded components. Increasing $\alpha$ to $1.8$ or $2.0$ produces similarly smooth boundaries, providing little additional structural benefit. We therefore use $\alpha=1.5$, $L=6$, and persistence $\rho=0.8$ for all eight MVTec AD 2 categories. The field is defined in normalized frequency coordinates, allowing the same configuration to be applied across categories and image resolutions without category-specific tuning.

\begin{figure}[t]
  \centering
  \includegraphics[width=0.99\columnwidth]{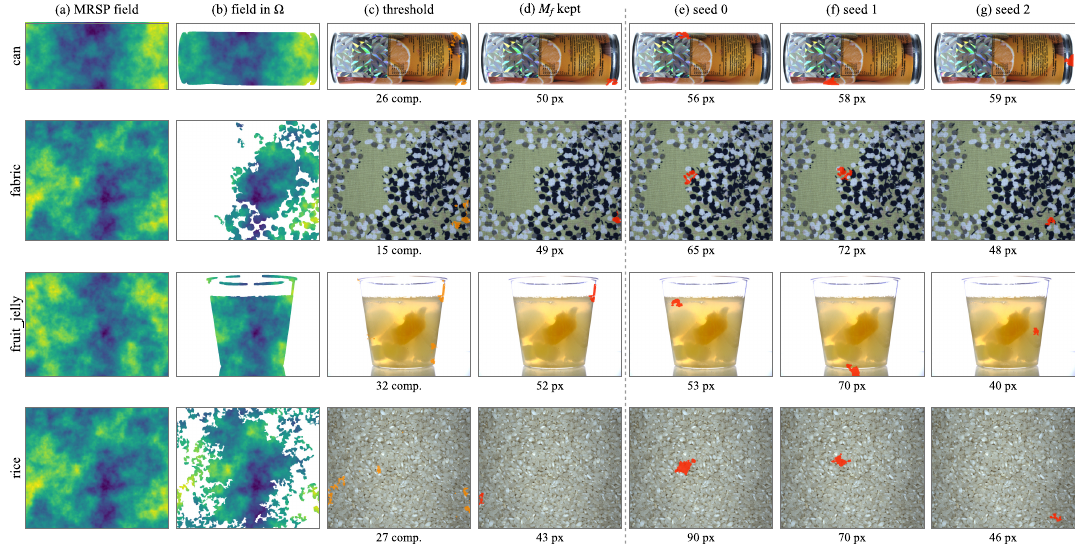}
  \caption{Construction of the object-aware placement mask $M_f$ for Can, Fabric, Fruit Jelly, and Rice. Each row shows the MRSP field, its restriction to the object region $\Omega$, thresholded response, retained largest connected component, and seed-dependent placement regions.}
  \label{fig:placement_mask_A}
\end{figure}

\begin{figure}[t]
  \centering
  \includegraphics[width=0.99\columnwidth]{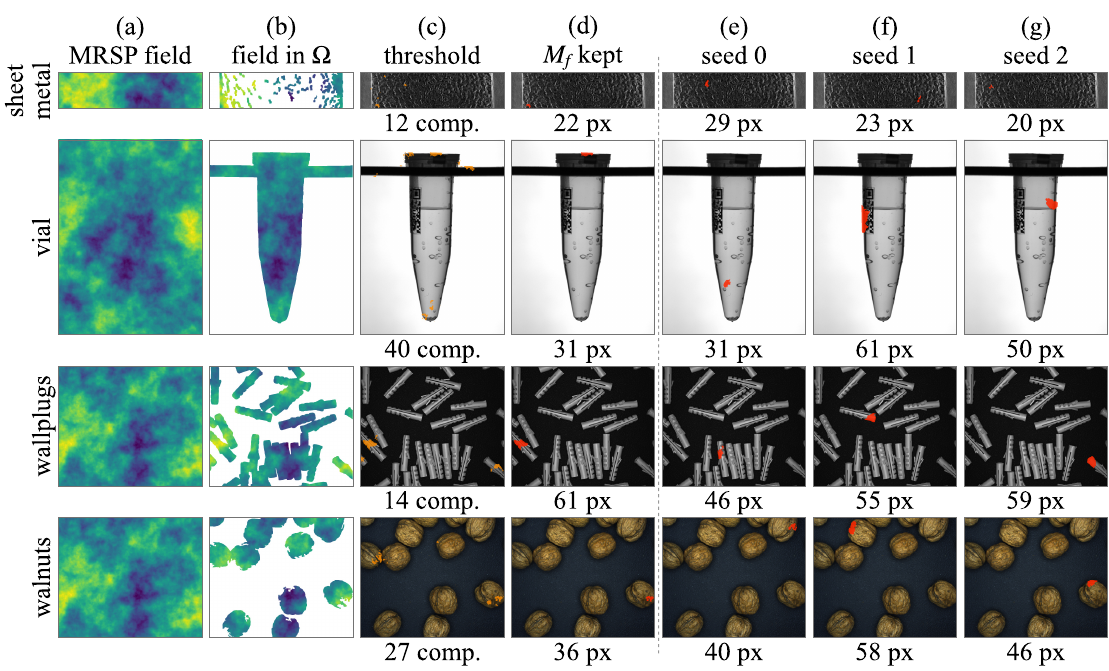}
  \caption{Construction of the object-aware placement mask $M_f$ for Sheet Metal, Vial, Wallplugs, and Walnuts. Each row shows the MRSP field, its restriction to the object region $\Omega$, thresholded response, retained largest connected component, and seed-dependent placement regions.}
  \label{fig:placement_mask_B}
\end{figure}

\paragraph{From field to placement mask}
Because MRSP is defined over the full image whereas valid placement must remain on the object, OBS is applied before thresholding. Let $\Omega$ denote the object region produced by OBS. The MRSP threshold is computed only within $\Omega$, and the largest connected component of the thresholded response is retained as the final placement mask $M_f$. Figures~\ref{fig:placement_mask_A} and~\ref{fig:placement_mask_B} illustrate this construction across all eight MVTec AD 2 categories, progressing from the MRSP field to the object-restricted field, thresholded response, and retained component.

Computing the threshold within $\Omega$ makes the target coverage relative to the visible object rather than the complete image, avoiding category-dependent effects caused by different object occupancy. The connected-component filtering further enforces a single coherent placement site, consistent with the single-defect synthesis protocol. Before filtering, the thresholded fields produce 12--40 disconnected components across the illustrated categories; retaining the largest component reduces these responses to a single placement region.

\paragraph{Seed-dependent diversity}
Changing the MRSP seed alters the location and extent of $M_f$ while keeping the host image and construction procedure fixed, as illustrated in the seed variations of Figures~\ref{fig:placement_mask_A} and~\ref{fig:placement_mask_B}. For example, the retained-region widths vary across seeds by $90/70/46$\,px for Rice, $65/72/48$\,px for Fabric, and $29/23/20$\,px for Sheet Metal. This variation is generated by the stochastic placement field rather than by fitting a defect-size or location prior to real anomalies. Since the retrieved defect is scaled relative to the sampled $M_f$, the same mechanism provides spatial and extent diversity without requiring real anomalous images during mask construction.

\paragraph{Per-stage timing for FLASH algorithm}
Table~\ref{tab:supp-timing} shows the per-stage timing for our method. Blending dominates the per-image cost: Poisson blending (36\%) and CIELAB color harmonization (24\%) together account for roughly 60\% of the 586~ms, with OBS foreground extraction adding a further 26\%. In contrast, the components that distinguish FLASH from ordinary copy-paste synthesis are cheap, with MRSP noise generation, mask computation, and defect placement together costing under 10\% of the total. The only generative step, defect generation, is paid once per category (about 0.4~s of bank extraction plus offline $I_a$ generation) and never repeated per image, which is what makes the ``generate once, synthesize many'' formulation practical.

\begin{table}[t]
\centering
\scriptsize
\begin{tabular}{lcc}
\toprule
\textbf{Stage} & \textbf{Time (ms)} & \textbf{Share (\%)} \\
\midrule
OBS foreground extraction & 154 & 26.3 \\
MRSP noise generation & 12 & 2.0 \\
Placement-mask computation & 37 & 6.3 \\
Defect placement & 5 & 0.9 \\
CIELAB color harmonization & 139 & 23.7 \\
Poisson blending & 212 & 36.2 \\
\midrule
Total (measured) & 586 & 100.0 \\
\bottomrule
\end{tabular}
\caption{Per-image timing breakdown of FLASH synthesis at 1024$\times$1024 resolution on a single NVIDIA RTX 3090. Stage times are means over fresh, non-cached images. The reported sub-stages sum to 559~ms, leaving 27~ms of the measured 586~ms total unattributed.}
\label{tab:supp-timing}
\end{table}

\subsection*{D. Calibration Analysis}
\label{sec:supp_calibration_analysis}

We evaluate synthetic anomalies for decision-threshold calibration across five anomaly-detection architectures: PaDiM, PatchCore, AnomalyDINO, Dinomaly, and SuperADD. For each detector, normal representations and memory banks are fitted once on normal training data and held fixed, ensuring that downstream performance variations stem exclusively from the calibration data source. We benchmark thresholds calibrated on FLASH against procedural synthesis (Perlin noise) and generative style-transfer (AnoStyler), using calibration on real test anomalies as the empirical oracle reference.

\begin{table}[t]
\centering
\scriptsize
\begin{tabular}{lcccc}
\toprule
\textbf{Category} & \textbf{Real} & \textbf{Perlin} & \textbf{AnoStyler} & \textbf{FLASH (Ours)} \\
\midrule
\multicolumn{5}{l}{\textbf{PaDiM}} \\
\midrule
Can & 71.49 & 44.60 & 64.67 & 68.93 \\
Fabric & 73.89 & 73.17 & 58.37 & 73.17 \\
Fruit Jelly & 88.27 & 69.37 & 87.24 & 85.71 \\
Rice & 81.31 & 81.08 & 6.27 & 81.08 \\
Sheet Metal & 89.64 & 88.24 & 65.13 & 88.24 \\
Vial & 86.10 & 48.79 & 78.46 & 85.71 \\
Wallplugs & 75.24 & 75.00 & 63.80 & 75.00 \\
Walnuts & 77.06 & 75.00 & 75.00 & 75.00 \\
\midrule
\textbf{Mean} & 80.37 & 69.41 & 62.37 & \textbf{79.11} \\
\midrule
\multicolumn{5}{l}{\textbf{PatchCore}} \\
\midrule
Can & 70.92 & 37.86 & 42.56 & 45.33 \\
Fabric & 79.47 & 73.17 & 16.33 & 73.17 \\
Fruit Jelly & 92.16 & 46.64 & 54.16 & 73.56 \\
Rice & 80.73 & 81.08 & 11.42 & 81.08 \\
Sheet Metal & 89.00 & 88.24 & 26.75 & 88.24 \\
Vial & 91.02 & 64.28 & 84.67 & 85.71 \\
Wallplugs & 74.53 & 75.00 & 32.53 & 75.00 \\
Walnuts & 81.88 & 73.76 & 75.71 & 75.00 \\
\midrule
\textbf{Mean} & 82.46 & 67.50 & 43.02 & \textbf{74.64} \\
\midrule
\multicolumn{5}{l}{\textbf{AnomalyDINO}} \\
\midrule
Can & 71.34 & 46.75 & 44.28 & 52.45 \\
Fabric & 74.44 & 73.17 & 24.40 & 73.17 \\
Fruit Jelly & 86.83 & 24.76 & 48.14 & 85.71 \\
Rice & 84.26 & 81.08 & 17.98 & 81.08 \\
Sheet Metal & 88.69 & 88.24 & 58.08 & 88.24 \\
Vial & 91.61 & 54.81 & 74.65 & 85.71 \\
Wallplugs & 75.07 & 54.98 & 55.17 & 75.00 \\
Walnuts & 79.49 & 73.70 & 76.11 & 75.00 \\
\midrule
\textbf{Mean} & 81.47 & 62.19 & 49.85 & \textbf{77.05} \\
\midrule
\multicolumn{5}{l}{\textbf{Dinomaly}} \\
\midrule
Can & 71.43 & 43.21 & 51.48 & 50.81 \\
Fabric & 77.42 & 73.17 & 29.18 & 73.17 \\
Fruit Jelly & 87.03 & 6.32 & 45.02 & 85.28 \\
Rice & 81.16 & 81.08 & 40.71 & 81.08 \\
Sheet Metal & 88.66 & 88.24 & 50.27 & 88.24 \\
Vial & 89.05 & 56.19 & 87.57 & 85.71 \\
Wallplugs & 76.27 & 56.15 & 55.82 & 75.63 \\
Walnuts & 83.43 & 79.02 & 82.94 & 80.15 \\
\midrule
\textbf{Mean} & 81.81 & 60.42 & 55.37 & \textbf{77.51} \\
\midrule
\multicolumn{5}{l}{\textbf{SuperADD}} \\
\midrule
Can & 71.19 & 61.14 & 61.46 & 61.10 \\
Fabric & 75.60 & 73.17 & 41.03 & 73.17 \\
Fruit Jelly & 85.86 & 80.74 & 81.54 & 85.71 \\
Rice & 86.60 & 81.08 & 25.17 & 81.08 \\
Sheet Metal & 89.75 & 88.24 & 70.47 & 88.24 \\
Vial & 99.52 & 98.61 & 99.84 & 85.71 \\
Wallplugs & 76.47 & 75.00 & 52.87 & 75.00 \\
Walnuts & 84.09 & 75.00 & 82.51 & 75.00 \\
\midrule
\textbf{Mean} & 83.64 & \textbf{79.12} & 64.36 & 78.13 \\
\bottomrule
\end{tabular}
\caption{Per-category image-level F1 (\%) across all five detectors on MVTec AD 2. The Mean row averages over the eight categories; bold marks the best synthetic source per detector (the Real oracle is excluded).}
\label{tab:supp-image-f1}
\end{table}

\begin{table}[t]
\centering
\scriptsize
\begin{tabular}{lcccc}
\toprule
\textbf{Category} & \textbf{Real} & \textbf{Perlin} & \textbf{AnoStyler} & \textbf{FLASH (Ours)} \\
\midrule
\multicolumn{5}{l}{\textbf{PaDiM}} \\
\midrule
Can & 0.16 & 0.05 & 0.05 & 0.05 \\
Fabric & 3.36 & 1.44 & 0.93 & 1.01 \\
Fruit Jelly & 12.41 & 4.46 & 5.13 & 2.96 \\
Rice & 6.26 & 2.03 & 6.21 & 1.74 \\
Sheet Metal & 11.95 & 3.70 & 11.27 & 9.08 \\
Vial & 10.01 & 0.18 & 3.22 & 0.37 \\
Wallplugs & 1.09 & 0.88 & 0.94 & 0.72 \\
Walnuts & 15.83 & 11.63 & 15.23 & 15.21 \\
\midrule
\textbf{Mean} & 7.63 & 3.05 & \textbf{5.37} & 3.89 \\
\midrule
\multicolumn{5}{l}{\textbf{PatchCore}} \\
\midrule
Can & 0.06 & 0.04 & 0.00 & 0.00 \\
Fabric & 15.27 & 1.88 & 2.05 & 15.19 \\
Fruit Jelly & 40.02 & 38.92 & 36.06 & 14.28 \\
Rice & 22.81 & 2.64 & 0.00 & 3.26 \\
Sheet Metal & 30.69 & 5.12 & 0.01 & 26.32 \\
Vial & 32.23 & 17.66 & 27.54 & 28.14 \\
Wallplugs & 16.22 & 2.76 & 15.94 & 8.53 \\
Walnuts & 51.41 & 50.15 & 47.04 & 49.38 \\
\midrule
\textbf{Mean} & 26.09 & 14.90 & 16.08 & \textbf{18.14} \\
\midrule
\multicolumn{5}{l}{\textbf{AnomalyDINO}} \\
\midrule
Can & 0.06 & 0.04 & 0.00 & 0.00 \\
Fabric & 46.11 & 16.61 & 20.11 & 37.66 \\
Fruit Jelly & 40.22 & 15.92 & 13.42 & 21.26 \\
Rice & 58.47 & 31.29 & 1.84 & 56.34 \\
Sheet Metal & 32.09 & 9.37 & 3.12 & 8.16 \\
Vial & 32.45 & 9.00 & 25.64 & 27.30 \\
Wallplugs & 2.39 & 1.57 & 0.00 & 1.83 \\
Walnuts & 56.59 & 55.95 & 47.82 & 56.30 \\
\midrule
\textbf{Mean} & 33.55 & 17.47 & 13.99 & \textbf{26.11} \\
\midrule
\multicolumn{5}{l}{\textbf{Dinomaly}} \\
\midrule
Can & 0.03 & 0.03 & 0.00 & 0.00 \\
Fabric & 27.45 & 15.68 & 3.77 & 19.46 \\
Fruit Jelly & 52.89 & 28.04 & 9.76 & 37.18 \\
Rice & 46.46 & 22.35 & 0.00 & 21.21 \\
Sheet Metal & 44.24 & 15.00 & 0.00 & 30.90 \\
Vial & 35.46 & 0.58 & 9.36 & 23.97 \\
Wallplugs & 2.36 & 1.27 & 0.00 & 2.18 \\
Walnuts & 46.27 & 40.70 & 3.90 & 45.35 \\
\midrule
\textbf{Mean} & 31.90 & 15.45 & 3.35 & \textbf{22.53} \\
\midrule
\multicolumn{5}{l}{\textbf{SuperADD}} \\
\midrule
Can & 0.02 & 0.01 & 0.00 & 0.00 \\
Fabric & 78.36 & 18.57 & 44.30 & 69.06 \\
Fruit Jelly & 56.07 & 40.02 & 55.93 & 55.57 \\
Rice & 58.75 & 6.21 & 25.36 & 51.90 \\
Sheet Metal & 35.87 & 2.53 & 7.79 & 7.39 \\
Vial & 57.05 & 56.46 & 46.63 & 39.41 \\
Wallplugs & 54.30 & 1.98 & 50.79 & 17.41 \\
Walnuts & 71.86 & 29.78 & 65.12 & 66.28 \\
\midrule
\textbf{Mean} & 51.53 & 19.44 & 36.99 & \textbf{38.38} \\
\bottomrule
\end{tabular}
\caption{Per-category pixel-level F1 (\%) across all five detectors on MVTec AD 2. The Mean row averages over the eight categories; bold marks the best synthetic source per detector (the Real oracle is excluded).}
\label{tab:supp-pixel-f1}
\end{table}

\noindent\textbf{Image-level calibration performance.}
Table~\ref{tab:supp-image-f1} provides the per-category image-level F1 scores across all detectors. FLASH yields the highest synthetic mean image-level F1 across four of the five architectures: 79.11\% for PaDiM (vs. 80.37\% oracle), 74.64\% for PatchCore (vs. 82.46\% oracle), 77.05\% for AnomalyDINO (vs. 81.47\% oracle), and 77.51\% for Dinomaly (vs. 81.81\% oracle). Across these detectors, FLASH recovers between 91\% and 98\% of the oracle performance.

Per-category image-level F1 varies across object geometries. FLASH reaches 73.17\% on \textit{fabric} and 81.08\% on \textit{rice} across four architectures. In contrast, AnoStyler obtains lower scores on categories such as \textit{rice} (6.27\% on PaDiM and 11.42\% on PatchCore) and \textit{fabric} (16.33\% on PatchCore), indicating weaker calibration transfer for these categories. While Perlin achieves a higher mean image-level F1 on SuperADD (79.12\% vs. 78.13\% for FLASH), this advantage does not carry over to mean pixel-level F1.

\noindent\textbf{Pixel-level localization calibration.}
Table~\ref{tab:supp-pixel-f1} presents the per-category pixel-level F1 scores. Pixel-level calibration is substantially more sensitive to synthetic artifact alignment, as threshold over- or underestimation directly degrades defect mask precision. FLASH achieves the highest overall synthetic mean on PatchCore (18.14\%), AnomalyDINO (26.11\%), Dinomaly (22.53\%), and SuperADD (38.38\%), outperforming both procedural and generative baselines.

The advantage of FLASH is particularly evident on modern foundation-model detectors. On AnomalyDINO, FLASH attains a mean pixel F1 of 26.11\% (approaching the 33.55\% oracle reference), whereas Perlin and AnoStyler degrade to 17.47\% and 13.99\%, respectively. Similarly, on Dinomaly, FLASH attains 22.53\% (vs. 31.90\% oracle), compared to 15.45\% for Perlin and 3.35\% for AnoStyler. AnoStyler often falls below 1.0\% on \textit{can}, \textit{rice}, and \textit{wallplugs}, indicating limited pixel-level calibration transfer in these settings.

\noindent\textbf{Category-specific calibration dynamics.}
Evaluating the per-category distributions reveals distinct structural behaviors across synthetic paradigms:
\begin{itemize}
    \item \textit{Complex texture and particulate regimes:} On granular substrates such as \textit{rice} and textured surfaces such as \textit{fabric} and \textit{walnuts}, FLASH provides robust calibration. For instance, on \textit{rice} using SuperADD, FLASH achieves 51.90\% pixel F1 compared to 6.21\% for Perlin and 25.36\% for AnoStyler.     Object Boundary Suppression (OBS) and Multi-Resolution Spectral Pyramid (MRSP) noise restrict placement to the estimated object region, supporting localized defects.
    \item \textit{Transparent and specular regimes:} Highly reflective or transparent objects present distinct challenges. On \textit{vial}, Perlin provides strong pixel calibration (56.46\% on SuperADD and 17.66\% on PatchCore) due to high contrast against uniform backlighting, whereas FLASH achieves 39.41\% and 28.14\%, respectively. On \textit{sheet metal}, which exhibits severe dark-field illumination variations, all synthetic methods encounter reduced calibration performance relative to the oracle (35.87\%), though FLASH maintains competitive transfer (7.39\% on SuperADD and 30.90\% on Dinomaly).
    \item \textit{Zero-oracle degeneracy:} For the \textit{can} category, the detector itself fails to segment real anomalies even under oracle calibration (0.02\% on SuperADD, 0.03\% on Dinomaly, and 0.06\% on PatchCore),     a challenge also reported for SuperADD \cite{ref190}. All synthetic calibration sources also attain near-zero performance, indicating that threshold transfer alone does not resolve this category.
\end{itemize}

\subsection*{E. Per-Category Synthesis Sheets}
\label{sec:supp_sheets}

\begin{figure*}[t]
    \centering
    \includegraphics[width=\textwidth]{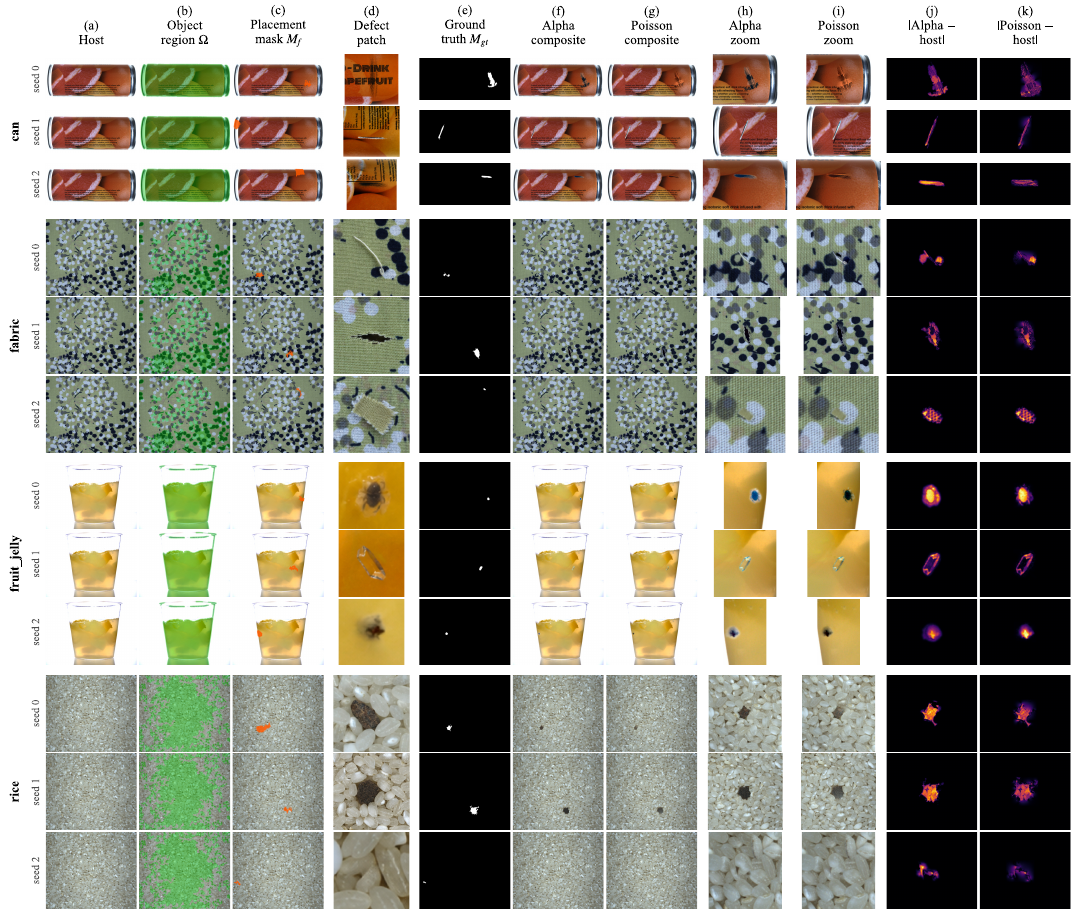}
    \caption{Per-category synthesis results for Can, Fabric, Fruit Jelly, and Rice. Each row traces a synthesized instance through the host image, object region $\Omega$, placement mask $M_f$, retrieved defect, ground-truth mask $M_{gt}$, the alpha and Poisson composites, their zoomed views, and the absolute difference of each composite against the host. Rows correspond to three seeds of the same category.}
    \label{fig:sheets_a}
\end{figure*}

\begin{figure*}[t]
    \centering
    \includegraphics[width=\textwidth]{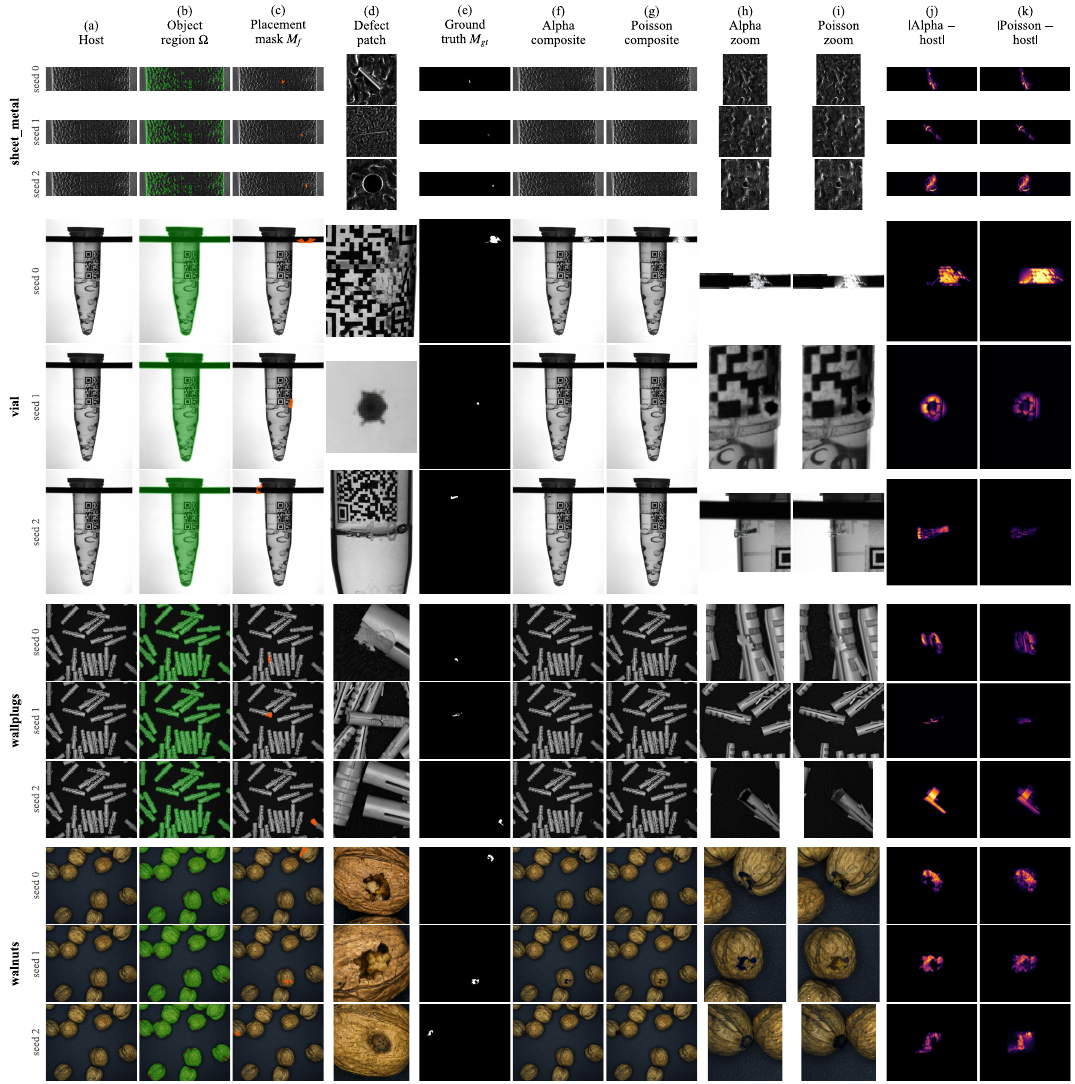}
    \caption{Per-category synthesis results for Sheet Metal, Vial, Wallplugs, and Walnuts; columns as in Figure~\ref{fig:sheets_a}. These categories carry the constrained supports: a thin specular strip, a transparent vial, and two multi-object scenes in which the placement mask selects which instance receives the defect.}
    \label{fig:sheets_b}
\end{figure*}

Figures~\ref{fig:sheets_a} and~\ref{fig:sheets_b} trace synthesis across all eight MVTec AD~2 categories. Each row shows the normal host, object-aware region $\Omega$, MRSP placement mask $M_f$, retrieved crop $C_d$, ground-truth mask $M_{gt}$, and the alpha and Poisson composites with their zoomed and difference views. Here, $M_f$ specifies the admissible placement region, whereas $M_{gt}$ records the transformed defect footprint. The category-level configuration remains fixed across seeds while placement and defect realization vary.

The harmonized crop is translated, rotated, scaled, and constrained by $M_f\cap\Omega$ before blending. Feathered alpha blending avoids the extra substrate collar required by Poisson blending, making it suitable for small placements. For larger defects, Poisson blending moves the boundary into the surrounding substrate to reduce visible transition artifacts. The paired outputs hold the host, defect, and placement fixed to expose these trade-offs.

Neither operator is assumed optimal at every scale. Alpha blending can retain contrast but leave a visible boundary, whereas Poisson blending can attenuate small defects. The placed defect's equivalent radius selects alpha blending for small defects and Poisson blending for larger ones; a failed Poisson solve falls back to alpha. This follows the adaptive synthesis procedure in the main paper.

The difference maps check whether changes remain near the defect support $M_{gt}$. Off-target changes would indicate compositing artifacts and weaken image-mask correspondence. Across the categories in Figures~\ref{fig:sheets_a} and~\ref{fig:sheets_b}, changes remain localized around the defects, with softer transitions in the Poisson outputs. These examples provide qualitative evidence of defect reuse while preserving host context.

\subsection*{F. Visual Comparison with Existing Synthetic Anomaly Generators}
\label{sec:supp_baselines}

\begin{figure}[t]
    \centering
    \includegraphics[width=0.99\columnwidth]{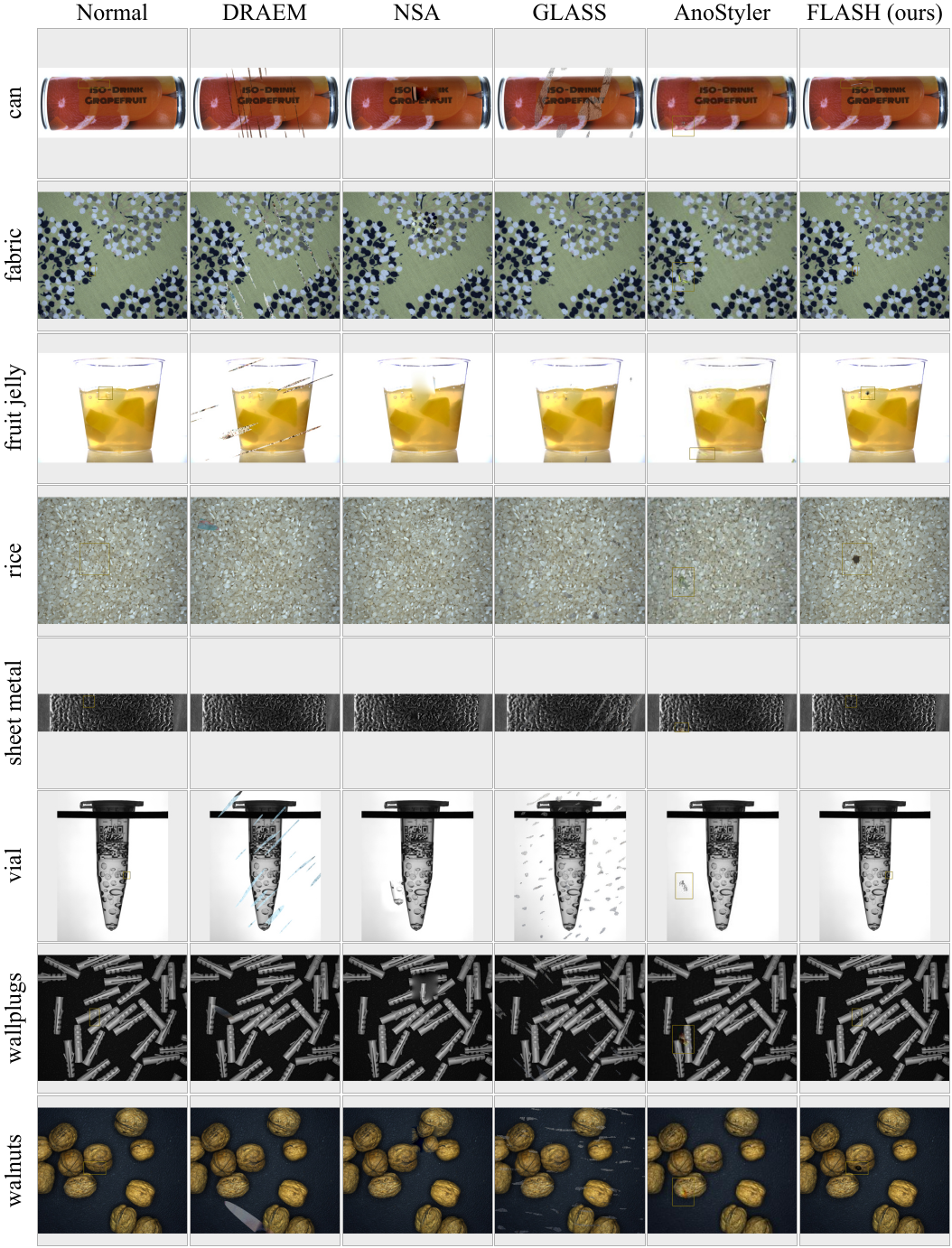}
    \caption{Visual comparison of synthetic anomaly generation across the eight MVTec AD~2 categories. Each row uses the same normal host image across methods, with columns showing the clean host followed by DRAEM, NSA, GLASS, AnoStyler, and FLASH (ours).}
    \label{fig:gallery}
\end{figure}

\paragraph{Comparison protocol}
We compare FLASH with DRAEM and GLASS as Perlin-based procedural generators, NSA as a same-category patch-transfer method with Poisson blending, and AnoStyler as a text-conditioned generative approach. These baselines represent the principal alternatives to the components of FLASH: procedural anomaly masks, seamless patch compositing, and semantic defect generation. All methods receive the same host images and are executed with a fixed seed through their native synthesis pipelines, without result selection. Figure~\ref{fig:gallery} shows one realization for each category and method.

\paragraph{Visual qualitative comparison}
DRAEM and GLASS use Perlin-based masks without explicit object-aware placement, leading to elongated or fragmented anomalies that can cross object boundaries or appear on the background, particularly for Can, Fruit Jelly, Vial, and Walnuts. DRAEM also produces substantially larger regions than typical localized defects. NSA provides material-consistent patches through same-category transfer and Poisson blending, but can introduce visible patch boundaries, weak low-frequency changes, or off-object placement. AnoStyler produces more plausible defect appearance in several cases, but semantic generation alone does not constrain the modification to the target object. In contrast, FLASH jointly specifies the defect semantically and constrains its placement using the object-aware mask, producing localized anomalies across all eight categories. This comparison is qualitative; a quantitative evaluation of the baseline generators as calibration sources under the same calibration protocol is left to future work.

\paragraph{Estimated generation cost}

Single-image generation measured on $2\times$\text{T4} GPUs requires approximately $52$\,ms for GLASS, $80$\,ms for NSA, $1.2$\,s for DRAEM, and $54$\,s for AnoStyler. Each value is a single call producing one image and its mask, and the reported figure is the median across categories after excluding the first category evaluated, which carries model load and process startup. These measurements include the execution overhead of each native pipeline: AnoStyler is driven as a separate process and reloads its diffusion and segmentation weights on every call, and the DRAEM wrapper constructs its texture-backed dataset per call, so both figures are upper bounds on marginal cost rather than marginal cost itself. Native-pipeline overhead and different hardware limit direct runtime comparisons. FLASH separates defect generation and extraction from subsequent synthesis: its estimated replay cost is $586.2$\,ms per image on an NVIDIA RTX\,3090, consistent with the estimate reported in the main paper, while the one-time defect-bank construction cost is amortized over subsequent samples. This structure allows a validated defect to be reused across multiple hosts and placements.

\subsection*{G. Ablation of the Defect Bank}
\label{sec:supp-bank}

\paragraph{Bank construction}
Stage~3 admits recovered defects into the defect bank $B$. Stage~5 draws its defects exclusively from $B$, so the admission criteria determine the set of defects available to every synthetic anomaly. A candidate is admitted when it passes a contrast test and a vision-language check. Candidates rejected by the vision-language check are subject to a visual review.

\emph{Candidate crops.} Each connected region of the DiffMask output with an area of at least $500$ pixels is treated as a candidate, so a single image may yield several. Each candidate is extracted as a square crop whose side equals $1.8$ times the longer side of the region's bounding box. The crop retains unchanged substrate around the defect, which the harmonization step requires. The mask is smoothed with a Gaussian of standard deviation $\sigma=1.0$ pixels to yield a soft alpha mask $\alpha_d$.

\emph{Contrast test.} A candidate passes if at least $10\%$ of its defect pixels ($\alpha_d>0.50$) differ by more than $\Delta E=6$ in CIELAB from the median color of the substrate pixels ($\alpha_d<0.10$). The substrate is taken from the crop itself, so the test is independent of the host image. The test rejects masks that cover unchanged material.

\emph{Vision-language check.} Each candidate that passes the contrast test is evaluated by Qwen2.5-VL-7B-Instruct under greedy decoding. The model receives the normal crop and the anomaly crop of the region, together with the question below. The category name is inserted with underscores replaced by spaces. A reply containing \texttt{"defect": true} admits the candidate; a reply that is not valid JSON is treated as a rejection.

\begin{quote}
\ttfamily\scriptsize
The first image is a normal region of a \{category\}. The second image is the same region after
an edit. Is there any visible change in the second image that a quality inspector would count as
a defect, contamination, damage, or a foreign object? Answer true for any such change, even small
or partial. Answer false only if the two images look the same. Reply only with JSON:
\{"defect": true or false, "reason": "short phrase"\}
\end{quote}

\emph{Visual review.} Candidates rejected by the vision-language check are presented to a reviewer with both crops, the mask, and the model's reply. A candidate is admitted if an anomaly is visible in the anomaly crop at the masked location, and rejected otherwise. Admitted candidates are flagged in the logs.

\emph{Storage.} Each admitted candidate is stored with its square RGB crop, its soft alpha mask $\alpha_d$, and its \texttt{defect\_frac}, defined as the longer side of the defect divided by the longer side of the frame. Stage~5 uses \texttt{defect\_frac} to set the size at which the defect is replayed. Entries are keyed \texttt{<donor\_id>\_<x>\_<y>}, where $(x,y)$ denotes the top-left corner of the region.

\paragraph{Effect of defect bank size}

We assess how the number of generated source images ($I_a$) used to construct the defect bank affects threshold transfer on three MVTec AD 2 categories: \texttt{rice}, \texttt{fruit jelly}, and \texttt{walnuts}. We denote this source-image count by $B_{\mathrm{size}}$ and evaluate $B_{\mathrm{size}}\in\{3,5,7,9,11,14\}$ with three fixed seeds. The three-image bank uses the original source images; larger banks are nested within each seed.
The number of synthetic calibration anomalies generated using FLASH is fixed at 90, 60, and 90
for rice, fruit jelly, and walnuts, respectively. As mentioned in Section~\ref{sec:experimental_setup}, image and pixel thresholds independently maximize calibration F1 on the generated synthetic anomalies. These thresholds are evaluated unchanged on the real public-test set of the  MVTec AD 2 dataset. For this ablation, we evaluate SuperADD, Dinomaly, PatchCore, and PaDiM. Dinomaly uses 1,000 optimizer steps and a
\texttt{392x392} center crop after resizing to \texttt{448x448} while the other detectors use
\texttt{448x448} inputs.
F1 scores use a 0--100 scale; all differences are absolute changes on this scale.

\begin{figure}[t]
    \centering
    \includegraphics[width=0.99\columnwidth]{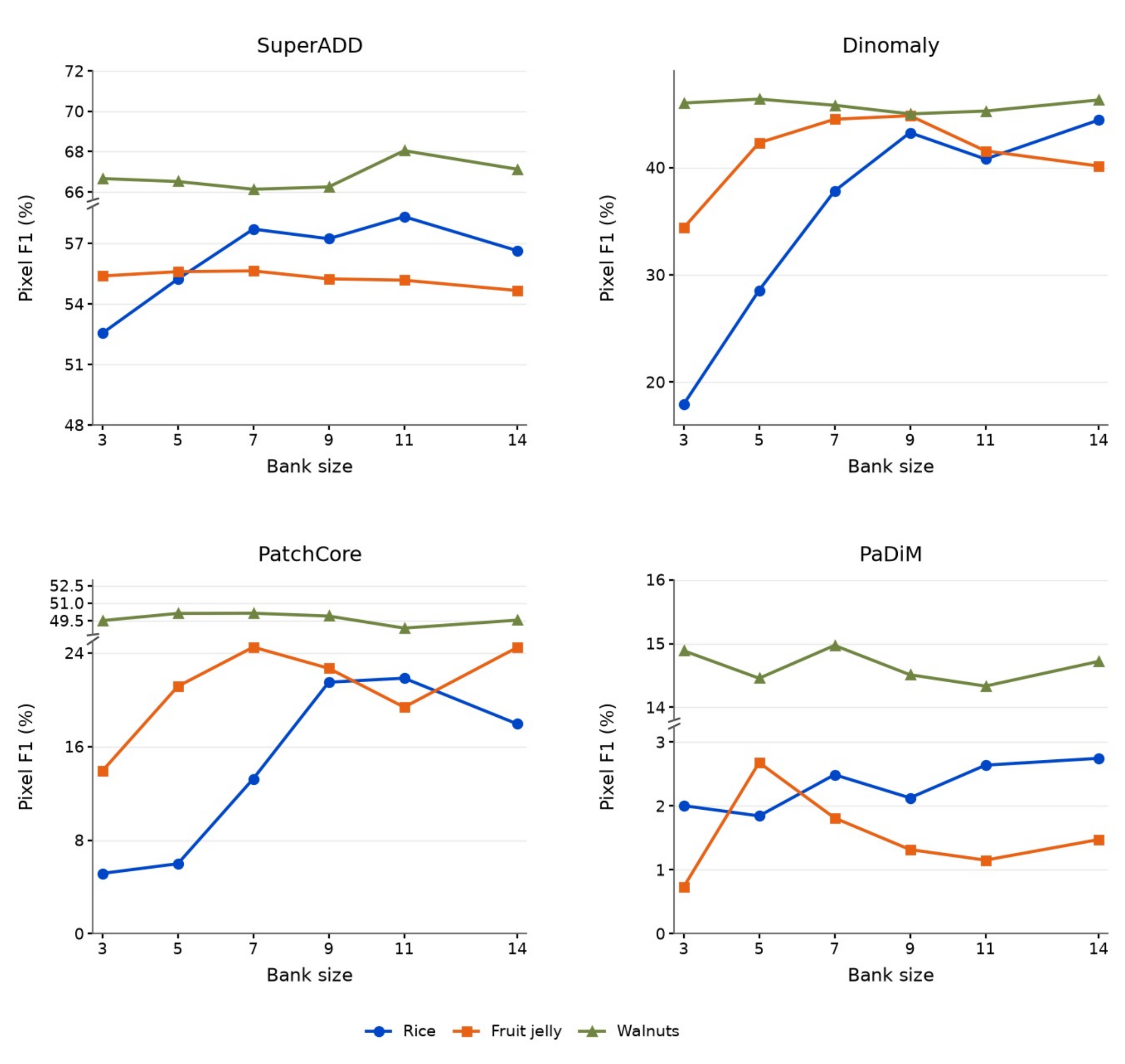}
    \caption{Effect of increasing defect bank size on pixel F1 across the anomaly detection models. Lines connect category-wise means over three seeds. Each plot uses a detector-specific y-axis range and axis breaks to best visualize the effect.}
    \label{fig:defect_bank_plots_suppl}
\end{figure}

\emph{Results}
Table~\ref{tab:defect_bank_size_ablation} shows pixel F1 averaged across the three categories. Figure~\ref{fig:defect_bank_plots_suppl} shows category-wise changes. From bank size 3 to 14, Dinomaly gains 10.85, followed by PatchCore (7.79), SuperADD (1.28), and PaDiM (0.44). Note that the three-image controls use the original source images; differences from Table~\ref{tab:supp-pixel-f1} may arise from fresh detector fits, explicit random-seed initialization, and changes to fit-time validation. All changes use the study-specific three-image controls.

\begin{table}[t]
    \centering
    \scriptsize
    \setlength{\tabcolsep}{4pt}
    \resizebox{\columnwidth}{!}{%
    \begin{tabular}{@{}rrrrr@{}}
        \toprule
        \textbf{Defect bank size} & \textbf{SuperADD} & \textbf{Dinomaly} & \textbf{PatchCore} & \textbf{PaDiM} \\
        \midrule
        3  & 58.20          & 32.77             & 22.86             & 5.88 \\
        5  & 59.12          & 39.08             & 25.75             & 6.33 \\
        7  & 59.83          & 42.71             & 29.30             & \textbf{6.42} \\
        9  & 59.58          & \textbf{44.36}    & \textbf{31.36 }   & 5.99 \\
        11 & \textbf{60.52} & 42.53             & 30.03             & 6.04 \\
        14 & 59.48          & 43.62             & 30.65             & 6.31 \\
        \bottomrule
    \end{tabular}
    }
    \caption{Defect bank size effect on pixel F1 (\%), averaged across three categories of MVTec AD 2 (\texttt{rice}, \texttt{fruit jelly}, and \texttt{walnuts}) and across 3 seeds.}
    \label{tab:defect_bank_size_ablation}
\end{table}

Gains are also category-dependent. From bank size 3 to 14, pixel F1 on \texttt{rice} improves for all four models, while \texttt{walnuts} changes little. The curves are non-monotonic. The highest observed macro pixel F1 occurs at $B_{\mathrm{size}}=9$ for Dinomaly (44.36\%) and PatchCore (31.36\%), $B_{\mathrm{size}}=11$ for SuperADD
(60.52\%), and $B_{\mathrm{size}}=7$ for PaDiM (6.42\%). These test-observed peaks do not establish optimal bank sizes, asymptotic saturation, or a universally applicable bank size.

\subsection*{H. Component Ablation and Category-wise Sensitivity}
\label{sec:flash_component_ablation}

\paragraph{Experimental protocol}

We evaluate how selected FLASH components affect threshold calibration.
We use SuperADD with the DINOv3 ViT-H/16+ backbone and $448\times448$ inputs on all eight MVTec AD 2 categories.
We compare six single-axis variants with the full-FLASH configuration. Each variant changes one synthesis setting while retaining the original three-source defect bank and the remaining generation parameters. This study is separate from the three-category bank-size ablation.

The variants test foreground restriction, the placement field, adaptive
placement, color harmonization, and blending. OBS off removes the foreground placement restriction but retains the host-eligibility policy. Color-only OBS retains the color cue and removes the texture cue.
The uniform-blob variant replaces MRSP with a compact, uniformly located blob of matched object-relative area.
Original placement only sets the adaptive-placement fraction to zero, instead of the default mixture of original and adaptive placement. The remaining variants disable CIELAB harmonization or replace hybrid
blending with alpha blending for every sample.

We follow the calibration protocol in Section~\ref{sec:experimental_setup}.
Detectors are fitted using normal training images, with native validation normals used for fit-time validation. For each category and each of the 3 seeds, we evaluate the ablated configurations using one shared fitted detector. The full-FLASH reference is fitted separately with the same seed. Baseline comparisons are therefore paired by category and seed, rather than by an identical fitted detector instance. For each configuration, we select separate image and pixel thresholds by maximizing F1 on native validation normals and synthetic anomalies. We keep the positive budget fixed across variants: 90 samples per category and
seed, except 60 for fruit jelly and 105 for vial. These budgets match the public-test anomaly counts, so the budget choice is test-informed. The thresholds are transferred unchanged to the real public-test set.
We average seeds within each category and then weight the eight categories equally.
F1 scores use a 0 - 100 scale; all differences are absolute changes on this scale. The component ablation uses PyTorch 2.14.0 (CUDA 13.0) and Anomalib 2.5.2-dev.0.

\paragraph{Aggregate component effects}

Table~\ref{tab:flash_component_macro} reports the aggregate results.
The re-evaluated full-FLASH reference obtains 38.61\% pixel F1 and
78.23\% image F1. Note that these values are from a separate evaluation and differ from
the main-paper results. We therefore compute all changes against this reference.

Removing the OBS restriction reduces pixel and image F1 by 1.71 and
4.62, respectively. Replacing MRSP with a matched-area uniform blob reduces the same scores
by 0.74 and 3.04. These comparisons support the full foreground restriction and structured
placement field at the aggregate level for this detector. Using original placement alone reduces pixel F1 by 1.07, while image F1 increases by 0.11. The mixed placement policy therefore has a pixel-level advantage in this
comparison, but not a consistent advantage across both metrics.

The remaining variants reveal different trade-offs. Disabling harmonization increases pixel F1 by 1.03.
Alpha-only blending increases it by 0.82. Neither changes the mean image F1. Color-only OBS gives the highest pixel F1 among the reported variants, 41.86\%, an increase of 3.24.
Thus, the full configuration is not the best observed setting for every calibration metric.

\begin{table}[t]
    \centering
    \scriptsize
    \setlength{\tabcolsep}{4pt}
    \resizebox{\columnwidth}{!}{%
    \begin{tabular}{@{}lrrrr@{}}
        \toprule
        \textbf{Configuration} & \textbf{Pixel F1 (\%)} &
        \textbf{$\Delta$ pixel F1} & \textbf{Image F1 (\%)} &
        \textbf{$\Delta$ image F1} \\
        \midrule
        Full FLASH              & $38.61 $ & $+0.00$ & $78.23 $ & $+0.00$ \\
        OBS off                 & $36.91 $ & $-1.71$ & $73.61 $ & $-4.62$ \\
        MRSP: uniform blob      & $37.88 $ & $-0.74$ & $75.19 $ & $-3.04$ \\
        Original placement only & $37.54 $ & $-1.07$ & $78.34 $ & $+0.11$ \\
        Harmonization off       & $39.64 $ & $+1.03$ & $78.23 $ & $+0.00$ \\
        Alpha-only blending     & $39.44 $ & $+0.82$ & $78.23 $ & $+0.00$ \\
        Color-only OBS         & $41.86 $ & $+3.24$ & $78.15 $ & $-0.07$ \\
        \bottomrule
    \end{tabular}
    }
    \caption{Single-axis component ablations on SuperADD.
    Scores are transferred-threshold real-test F1 (\%), macro-averaged over
    all eight MVTec AD 2 categories. Differences use the current full-FLASH control and are computed before
    rounding.}
    \label{tab:flash_component_macro}
\end{table}

\paragraph{Category-wise effects}

Table~\ref{tab:flash_component_category} presents category-wise pixel F1 changes for the same configurations. This analysis reveals opposing effects that are obscured by the macro average. For example, removing OBS reduces pixel F1 by 24.88 on fruit jelly and 23.17 on vial. It increases the score by 22.50 on sheet metal and 27.60 on wallplugs. These changes partly cancel, leaving a macro decrease of only 1.71. OBS is therefore influential, but its effect is not uniformly positive across categories.

The improvements from simpler settings are also category-dependent. Color-only OBS gains 18.04 on sheet metal but loses 8.31 on wallplugs and 2.06 on walnuts. Alpha-only blending gains 8.26 on vial but loses 4.57 on walnuts. These results do not identify a universally preferred cue or compositing operator. Calibration F1 also does not establish comparative visual quality. The alpha-only comparison changes the compositing policy and therefore cannot establish that the Poisson collar is unnecessary.

\begin{table*}[t]
    \centering
    \footnotesize
    \setlength{\tabcolsep}{4pt}
    \begin{tabular}{lrrrrrrrrr}
        \toprule
        \textbf{Configuration} & \textbf{Can} & \textbf{Fabric} &
        \textbf{\shortstack{Fruit\\Jelly}} & \textbf{Rice} &
        \textbf{\shortstack{Sheet\\Metal}} & \textbf{Vial} &
        \textbf{Wallplugs} & \textbf{Walnuts} & \textbf{Macro} \\
        \midrule
        Full FLASH (F1, \%) & $0.00$ & $69.12$ & $55.39$ & $52.55$ &
        $7.76$ & $38.91$ & $18.51$ & $66.67$ & $38.61$ \\
        \midrule
        OBS off & $+0.00$ & $-16.44$ & $-24.88$ & $-2.78$ &
        $+22.50$ & $-23.17$ & $+27.60$ & $+3.51$ & $-1.71$ \\
        MRSP: uniform blob & $+0.00$ & $-13.73$ & $-0.39$ & $-7.26$ &
        $+5.98$ & $+2.60$ & $+5.60$ & $+1.29$ & $-0.74$ \\
        Original placement only & $+0.00$ & $+5.20$ & $+0.60$ & $-3.89$ &
        $-0.06$ & $-8.23$ & $-2.31$ & $+0.10$ & $-1.07$ \\
        Harmonization off & $+0.00$ & $-0.94$ & $+0.47$ & $+4.38$ &
        $-0.14$ & $+4.22$ & $-0.37$ & $+0.61$ & $+1.03$ \\
        Alpha-only blending & $+0.00$ & $+5.38$ & $-0.34$ & $-0.14$ &
        $-0.47$ & $+8.26$ & $-1.54$ & $-4.57$ & $+0.82$ \\
        Color-only OBS & $+0.00$ & $+8.83$ & $+0.40$ & $+5.88$ &
        $+18.04$ & $+3.17$ & $-8.31$ & $-2.06$ & $+3.24$ \\
        \bottomrule
    \end{tabular}
    \caption{Category-wise pixel F1 effects for the configurations in
    Table~\ref{tab:flash_component_macro}.
    The first row gives absolute full-FLASH pixel F1 (\%).
    All subsequent rows give changes from that category's control averaged over three seeds.
    Positive values indicate higher F1.
    The macro column weights all eight categories equally.}
    \label{tab:flash_component_category}
\end{table*}

\paragraph{Interpretation and limitations}

The results indicate component sensitivity rather than a uniform benefit from every stage. Foreground restriction and structured placement improve aggregate F1 in the reported comparisons, while harmonization and hybrid blending may not be required to obtain the highest observed pixel F1. The category-wise reversals do not support a universally preferred configuration. The observed spread does not isolate synthesis uncertainty or establish statistical significance. The comparison is descriptive rather than an independent validation of a revised configuration. It does not isolate component interactions or establish transfer to other detectors. Further studies are needed to determine which components transfer across datasets and defect types.

\end{document}